\documentclass{article} 
\usepackage{main,times}

\usepackage{amsmath,amsfonts,bm}

\def\eqref#1{equation~\ref{#1}}

\def\1{\bm{1}}

\DeclareMathAlphabet{\mathsfit}{\encodingdefault}{\sfdefault}{m}{sl}
\SetMathAlphabet{\mathsfit}{bold}{\encodingdefault}{\sfdefault}{bx}{n}

\usepackage{hyperref}
\usepackage{url}
\usepackage{xurl}

\usepackage{xspace}
\usepackage{enumitem}
\usepackage{array,longtable}
\usepackage{booktabs}
\usepackage{algorithm}
\usepackage{algorithmic}
\usepackage{graphicx}
\usepackage[table]{xcolor}
\usepackage{tcolorbox}
\tcbuselibrary{breakable,skins}

\definecolor{promptBgSystem}{HTML}{f3f3f3}
\definecolor{promptBgTitle}{HTML}{e0efff}
\definecolor{ariseTitleBg}{HTML}{E6F6FF}
\definecolor{ariseTitleBlue}{HTML}{0369FF}

\makeatletter
\newenvironment{arisefrontmatter}{%
  \begin{tcolorbox}[
    enhanced, frame hidden, colback=ariseTitleBg, arc=10pt,
    boxrule=0pt, boxsep=0pt, left=12pt, right=12pt,
    top=16pt, bottom=12pt, before skip=0pt, after skip=12pt
  ]
  \renewcommand{\@maketitle}{%
    {\LARGE\bfseries\raggedright\@title\par}
    \medskip
    {\small\raggedright\@author\par}
    \global\let\arise@authornotes\@thanks
    \global\let\arise@authorfootnotetext\@footnotetext
    \global\let\arise@authorfntext\@makefntext
    \global\let\@thanks\@empty
  }
  \renewenvironment{abstract}{\par\medskip\normalsize}{\par}
}{%
  \medskip
  \noindent
  \begin{minipage}[b]{\dimexpr\linewidth-2.2cm\relax}
    \footnotesize
    \textbf{Project page}\\
    \url{https://foundation-model-research.github.io/ARISE}
  \end{minipage}\hfill
  \includegraphics[width=1.8cm]{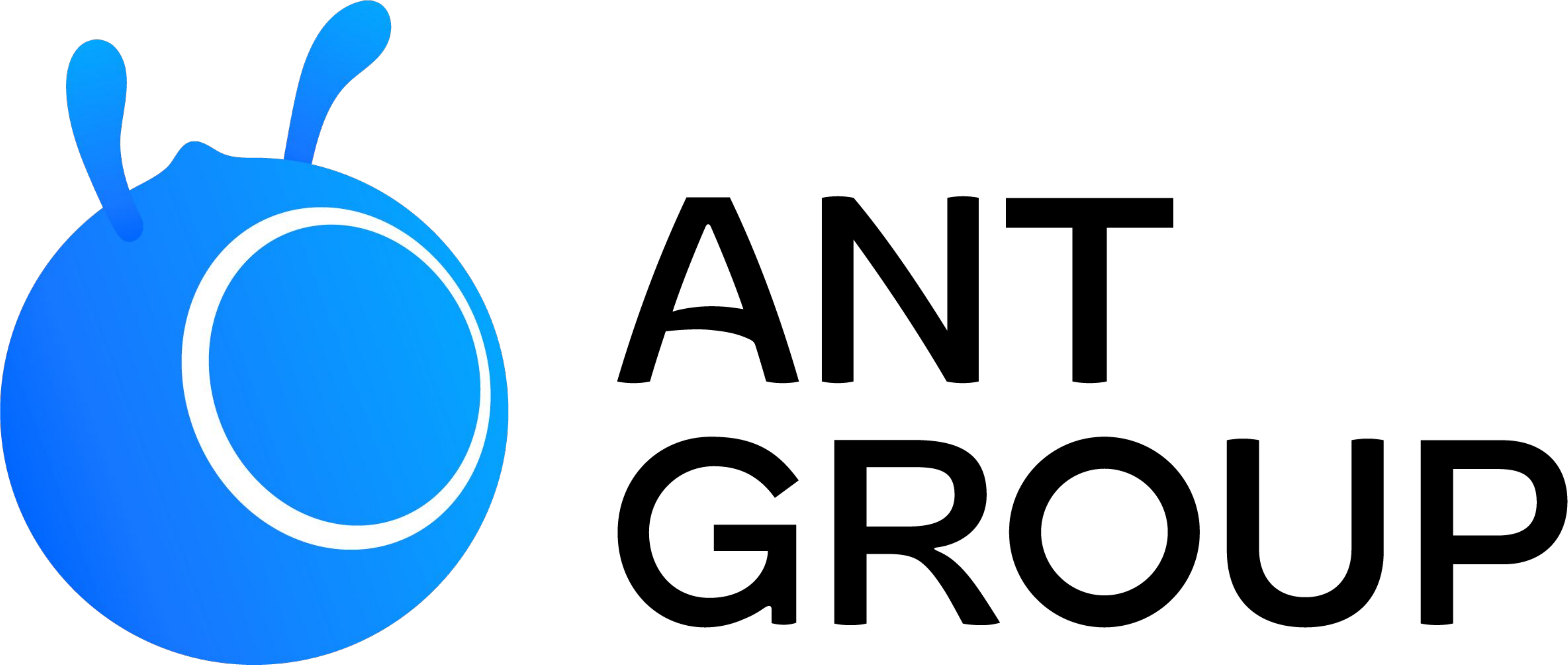}
  \end{tcolorbox}
  \begingroup
    
    \let\@footnotetext\arise@authorfootnotetext
    \let\@makefntext\arise@authorfntext
    \arise@authornotes
  \endgroup
  \setcounter{footnote}{0}
  \global\let\arise@authornotes\@empty
}
\makeatother

\newcommand{\method}{\textsc{Arise}\xspace}
\newcommand{\minisection}[1]{\vspace{5pt}\noindent\textbf{#1.}}

\title{\textcolor{ariseTitleBlue}{\method}: Adapting to Evolving Capability Gaps in Agentic Reinforcement Learning}

\author{
\textbf{Kun Feng$^{1, 2}$\thanks{
Equal contribution.
Kun Feng conducted this work during an internship at Ant Group.
} ,
Yuchen Fang$^{2}$\footnotemark[1] ,
Yiyang Tan$^{1}$,
Shuqi Gu$^{1}$,
Yongxiang Zhao$^{1}$}
\\
\textbf{
Yu Liu$^{2}$,
Xingyu Lu$^{2}$,
Lintao Ma$^{2}$,}
\textbf{Kan Ren$^{1}$\thanks{Corresponding author: \texttt{renkan@shanghaitech.edu.cn}}} \\
$^{1}$ShanghaiTech University \quad $^{2}$Ant Group \\
\texttt{\{fengkun2025,renkan\}@shanghaitech.edu.cn}, \texttt{yuchen.fyc@antgroup.com}
}

\iclrfinalcopy

\begin{document}
\pagestyle{plain}
\begin{arisefrontmatter}
\maketitle

\begin{abstract}
As a long-horizon agent improves through experience, previously observed weaknesses may recede while new limitations emerge, continually changing what it still needs to learn.
Yet the learning process often remains tied to a static view of these needs: fixed behavioral criteria and training priorities can become misaligned with evolving agent capabilities, while sparse task-level feedback makes such misalignment more difficult to detect.
Even when capability gaps are identified, rollouts from the current policy may repeatedly reproduce the same failures rather than explore better alternatives.
To address this, we introduce \textbf{A}daptive \textbf{R}ubr\textbf{i}c--\textbf{S}kill Co-\textbf{E}volution (\method), a reinforcement learning framework that uses rollout evidence to continually adapt evaluation criteria, exploration guidance, and training priorities.
Rubrics evolve to reward partial behavioral progress, while their paired skills are refined and selectively activated to guide exploration toward unresolved weaknesses.
Alongside this co-evolution, capability-based adaptive sampling prioritizes tasks that target behaviors needing further improvement.
Experiments on two challenging long-horizon agent benchmarks, SkillsBench and Terminal-Bench, demonstrate that \method successfully enhances both overall task performance and training efficiency.
\end{abstract}
\end{arisefrontmatter}

\section{Introduction}
Large Language Models (LLMs) increasingly act as agents that reason, invoke tools, and interact with environments to complete multi-step tasks~\citep{yao2023react,yang2024sweagent,feng2026kairosagent}.
These long-horizon tasks require the coordination of capabilities such as planning, execution, and verification~\citep{yao2023react,shinn2023reflexion}.
Reinforcement learning refines these behaviors through interactive experience~\citep{ivison2026tmax}.
Although recent methods enrich this process with behavioral feedback and reusable guidance~\citep{chen2026rucl,shao2026dr,xia2026skillrl}, structuring training for sustained capability development remains an open question.

Two coupled challenges hinder aligning training with evolving agent capabilities:
(i) \textit{Tracking evolving capability gaps.}
Correcting an intermediate error may leave the task reward unchanged if subsequent steps prevent completion, masking shifts in behavioral bottlenecks~\citep{lightman2023verify}.
Furthermore, even detailed criteria become uninformative once consistently satisfied, while out-of-scope bottlenecks remain unmeasured.
(ii) \textit{Exploring beyond repeated failures.}
While precise evaluation distinguishes sampled behaviors, it cannot guarantee the generation of useful alternatives.
Under group-relative optimization, identical task rewards within a rollout group yield zero advantage, even if trajectories differ in intermediate steps~\citep{shao2024deepseekmath,yu2025dapo}.
Consequently, mastered or excessively difficult tasks waste rollout budgets without providing useful reward contrasts, making their learning value highly dependent on the policy's developmental stage~\citep{florensa2018goalgan}.
Together, these challenges necessitate training paradigms that continually identify areas for improvement and foster the exploration and reinforcement of corresponding behaviors.

\begin{figure}[t]
  \centering
  \includegraphics[width=\linewidth]{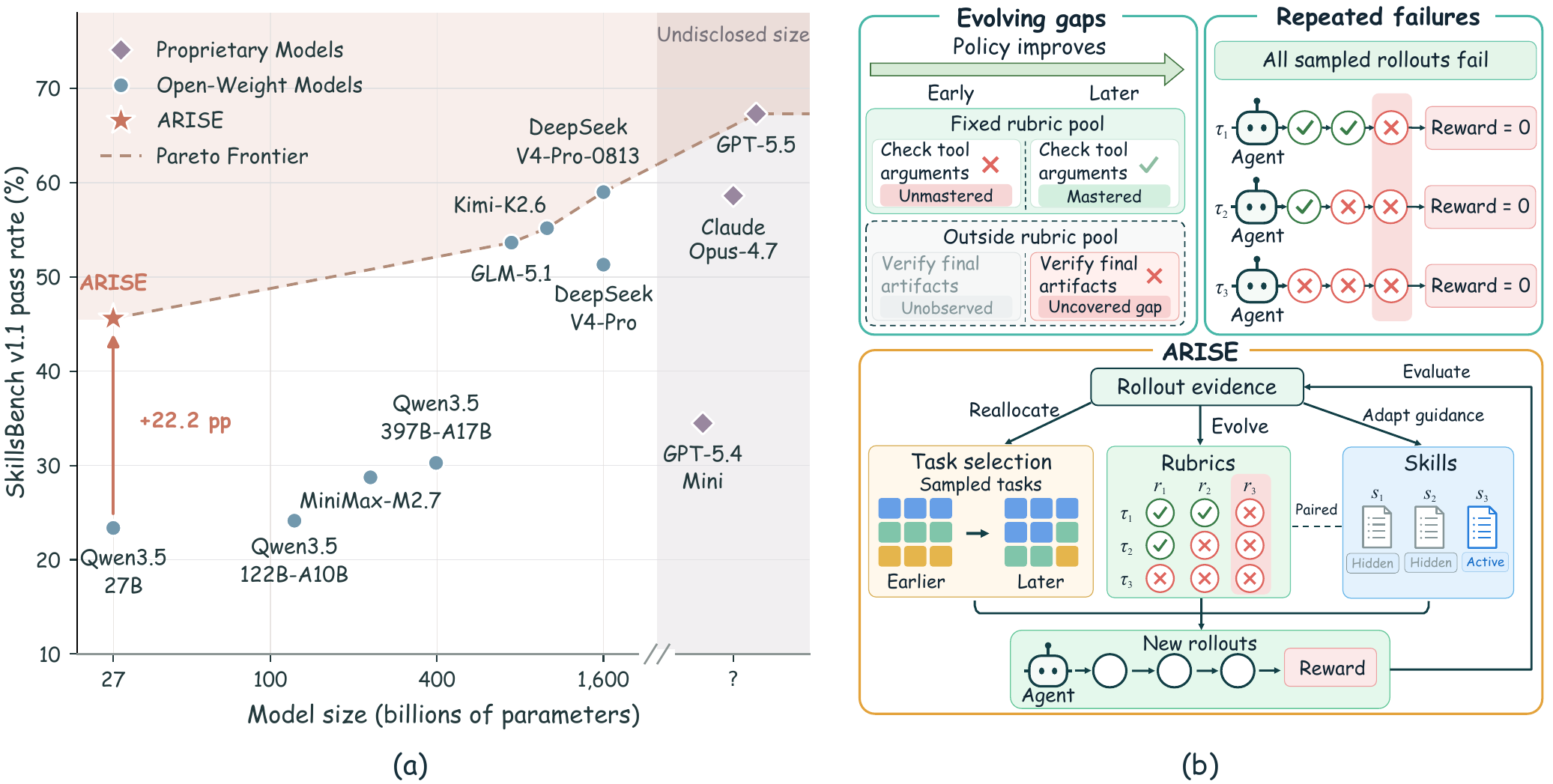}
  \vspace{-10pt}
  \caption{Performance and motivation of \method.
  (a) SkillsBench v1.1 pass rates versus model size, highlighting the improvement over Qwen3.5-27B and competitive performance with substantially larger models.
  (b) Fixed criteria can miss emerging capability gaps, while sampled rollouts may all fail despite making partial progress.
  \method uses rollout evidence to adapt rubrics, skill guidance, and task selection as agent capabilities evolve.
  }
  \vspace{-5pt}
  \label{fig:intro}
\end{figure}

To address these challenges, we propose \textbf{A}daptive \textbf{R}ubr\textbf{i}c--\textbf{S}kill Co-\textbf{E}volution (\method), a reinforcement learning framework driven by evolving capability gaps (Figure~\ref{fig:intro}(b)).
It translates rollout evidence of these gaps into concrete behavioral requirements, each formalized as a rubric for evaluating partial progress and a paired skill for guiding action. 
Rubric pass rates then direct exploration: low rates trigger skill activation or refinement, whereas consistently high rates indicate mastery, allowing the criterion to be retired.
As new evidence emerges, the framework expands this repertoire to cover unaddressed behaviors, ensuring the requirements remain responsive to policy development.

Progress on these requirements also depends on task context, as a behavior performed reliably in one task type may remain difficult in another.
We therefore propose a capability-based adaptive sampler that guides task selection using rubric-based estimates of capability performance within each task type.
Subsequent rollouts provide feedback for updating rubrics, paired skills, and the sampling distribution, ensuring that evolving capability assessments shape future training.
As shown in Figure~\ref{fig:intro}(a), \method significantly improves SkillsBench performance over its base model, achieving competitive results with much larger models.

In summary, our contributions are three-fold:
\begin{itemize}[leftmargin=3mm]
  \item We identify evolving capability gaps as a central challenge in long-horizon agent learning. Instead of relying on static behavioral requirements or training priorities, \method continuously leverages rollout evidence to dynamically uncover and address emerging bottlenecks.
  \item We propose \textbf{A}daptive \textbf{R}ubr\textbf{i}c--\textbf{S}kill Co-\textbf{E}volution (\method), which couples capability identification, targeted exploration, and fine-grained behavioral feedback, together with capability-based adaptive sampling.
  More broadly, it establishes a general paradigm for aligning the learning process with evolving agent capabilities.
  \item Empirically, we demonstrate that adapting training to these evolving capabilities yields substantial and efficient performance gains. Across two agent benchmarks, \method consistently outperforms same-scale baselines, proving to be a highly effective alternative to simply scaling model size.
\end{itemize}

\section{Related Work}

\paragraph{Rubric-Based Reinforcement Learning.}
Rubric-based approaches deconstruct evaluation into explicit criteria, aggregating individual judgments to derive training rewards~\citep{gunjal2025rubrics,viswanathan2025checklists}.
To make these evaluations adaptive, frameworks like OnlineRubrics~\citep{rezaei2025online} extract criteria by comparing current and reference outputs, while DR Tulu~\citep{shao2026dr} continuously updates rubrics based on on-policy responses and search contexts.
However, these response-level evaluations often overlook improvements in intermediate actions that are not reflected in the final output.
RuscaRL~\citep{zhou2025ruscarl} further uses rubrics as both rewards and exploration scaffolds, yet relies on fixed underlying criteria and decays guidance according to a predefined schedule.
In contrast, \method evaluates intermediate agent behaviors and dynamically revises active criteria as rollout evidence reveals new or consistently mastered behavioral requirements.

\paragraph{Skill Learning for LLM-Based Agents.}
Skill learning enables LLM-based agents to reuse experience through executable routines or natural-language guidance~\citep{wang2024agent,cai2024large}.
Voyager~\citep{wang2023voyager} builds a library of executable skills via environment interaction, while ExpeL~\citep{zhao2023expel} distills transferable insights.
By keeping policy parameters frozen, both methods rely entirely on the in-context capabilities of the base model to leverage this accumulated knowledge.
SkillRL~\citep{xia2026skillrl} extends experience reuse to reinforcement learning through a hierarchical skill library, evolving it based on validation success rates.
It consistently includes general skills while retrieving task-specific ones via semantic relevance.
However, relevance alone does not necessitate guidance, as a retrieved skill may describe behaviors the policy has already mastered.
In \method, evaluations of paired behavioral criteria determine exactly when guidance is needed, providing actionable evidence to refine skills if specific weaknesses persist.

\paragraph{Adaptive Task Sampling.}
Adaptive task sampling allocates training experience according to the evolving learning needs of the policy~\citep{jiang2021prioritized}.
GoalGAN~\citep{florensa2018goalgan} selects goals of intermediate difficulty based on empirical success rates.
In reinforcement learning for LLMs, 
DAPO~\citep{yu2025dapo} filters groups post-rollout to exclude those lacking reward variation, while VADE~\citep{hu2025vade} selects informative samples pre-rollout using online correctness estimates.
However, these outcome-based signals capture overall task difficulty without distinguishing the underlying behavioral causes: tasks with similar success rates may require improvements in entirely different capabilities.
The distinction in \method lies in the evidence used for selection: rubric evaluations expose which behaviors remain weak within each task type, rather than only whether tasks succeed.

\section{Methodology}

\subsection{Problem Formulation}
We consider an LLM-based agent that interacts with an environment to complete a task $x$ drawn from a task distribution $\mathcal{D}$.
Each task specifies an instruction, an execution environment, and the available tools.
At interaction step $t$, the agent observes a history $h_t=(x,a_1,o_1,\ldots,a_{t-1},o_{t-1})$ and samples an action $a_t\sim\pi_\theta(\cdot\mid h_t)$, where $\pi_\theta$ is the parameterized policy.
An action is a textual response or a tool invocation, and $o_t$ denotes the environment feedback resulting from $a_t$.
The interaction terminates after $T$ steps, producing a trajectory $\tau=(a_1,o_1,\ldots,a_T,o_T)$.

A task-specific verifier assigns an outcome reward $R_{\mathrm{task}}(x,\tau)$ that measures task completion.
The learning objective is to improve expected task performance:
\begin{equation}
    \max_\theta\; J(\theta)
    = \mathbb{E}_{x\sim\mathcal{D}}
      \mathbb{E}_{\tau\sim P_\theta(\cdot\mid x)}
      \left[R_{\mathrm{task}}(x,\tau)\right],
    \label{eq:task_objective}
\end{equation}
where $P_\theta(\cdot\mid x)$ is the trajectory distribution induced by the policy and environment.

\subsection{Overall Framework}
\begin{figure}[t]
  \centering
  \includegraphics[width=\linewidth]{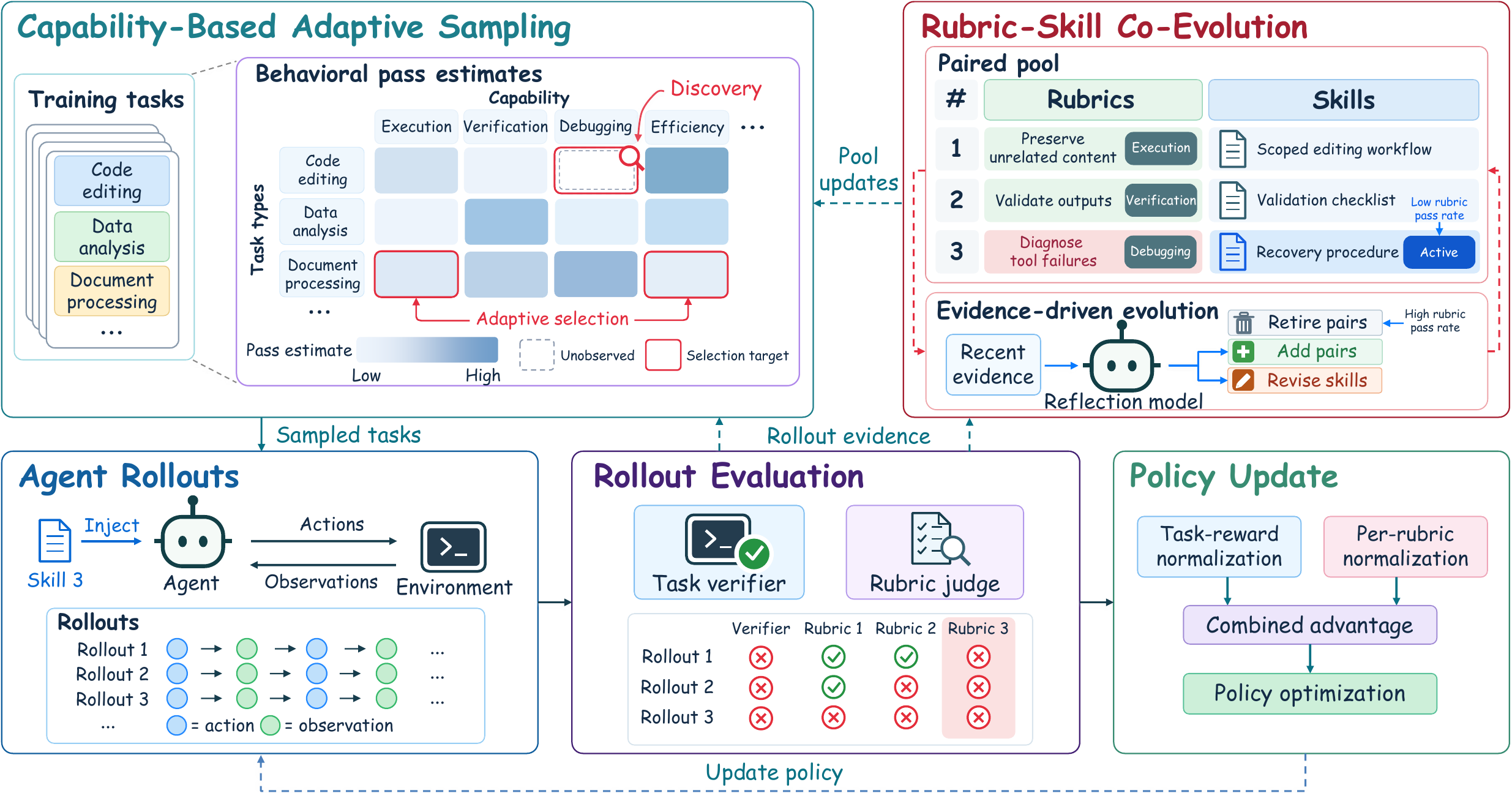}
    \caption{Overview of \method. Tasks sampled via capability-based priorities generate rollouts with selectively activated skill guidance. Rubric evaluations provide behavioral feedback for policy optimization and, alongside rollout evidence, drive rubric--skill evolution and subsequent task sampling.}
    \label{fig:framework}
\end{figure}

To track evolving capability gaps and support exploration beyond repeated failures, \method combines two components: \textit{rubric--skill co-evolution} and \textit{capability-based adaptive sampling}.
Together, they determine what behavioral feedback and guidance to provide and which tasks to train on.
Figure~\ref{fig:framework} presents the overall framework, and Appendix~\ref{app:training_procedure} summarizes the training procedure.

\subsubsection{Rubric--Skill Co-Evolution}
Rubric--skill co-evolution adapts what the agent is evaluated on and which guidance it receives as behavioral gaps change during training.
Prompt templates are provided in Appendix~\ref{app:prompts}.

\paragraph{Behavioral Criteria.}
Identifying a behavioral gap does not itself specify how to overcome it.
Each criterion therefore pairs an evaluative rubric $r_i$ with a skill $s_i$ targeting the same observable requirement.
The rubric defines the applicability conditions and the evidence required for satisfaction, while the skill translates this requirement into reusable natural-language action guidance.
A judge model evaluates applicable rubrics against the full interaction trajectory, including agent actions and environment feedback, rather than relying solely on the final response.
For each rubric $r_i$, we estimate its pass rate as:
\begin{equation}
    \hat p_i = \frac{1}{N_i}\sum_{j\in\mathcal{J}_i} y_{ij},
    \label{eq:rubric_pass_rate}
\end{equation}
where $\mathcal{J}_i$ indexes the trajectories to which rubric $r_i$ applies, $N_i=|\mathcal{J}_i|>0$, and $y_{ij}\in\{0,1\}$ represents the binary judge verdict (pass or fail).
The resulting pass rates guide skill activation, hiding, and refinement, as well as rubric retirement.

\paragraph{Evidence-Driven Evolution.}
As the policy improves, existing criteria may offer diminishing learning value or fail to capture emerging behavioral gaps.
The rubric--skill pool therefore retires consistently satisfied criteria and adds new pairs based on recent rollouts.
Initially, the pool is empty and training uses only task rewards.
A reflection model uses initial rollouts to generate the first rubric--skill pairs, enabling behavioral feedback and skill activation.
This reflection model periodically contrasts recent trajectories to identify emerging bottlenecks, consulting the current pool to avoid generating redundant criteria.
At training iteration $k\in\{0,1,\ldots\}$, the paired pool $\mathcal{P}_k$ consists of active rubrics $\mathcal{R}_k$ and their corresponding skills $\mathcal{S}_k$:
\begin{equation}
    \mathcal{P}_k
    = \{(r_i,s_i)\mid r_i\in\mathcal{R}_k,\; s_i\in\mathcal{S}_k\}.
    \label{eq:paired_pool}
\end{equation}
Rubric--skill lifecycle updates occur every $\Delta_{\mathrm{pool}}$ training iterations.
At each such update, new rubric--skill pairs are generated as:
\begin{equation}
    \mathcal{P}_k^{\mathrm{new}}
    = G_{\mathrm{ref}}\!\left(\mathcal{B}_k,\mathcal{P}_k\right),
    \qquad \Delta_{\mathrm{pool}}\mid(k+1),
    \label{eq:criterion_generation}
\end{equation}
where $G_{\mathrm{ref}}$ denotes the reflection model, and $\mathcal{B}_k$ contains structured evidence from recent rollouts, including tool-grounded behavioral diagnostics, environment feedback, skill-use summaries, and task contexts (Appendix~\ref{app:rollout_evidence}).
Each pair in $\mathcal{P}_k^{\mathrm{new}}$ encapsulates a reusable behavioral requirement, comprising an evaluative rubric and an actionable skill.
During scheduled updates, pairs are retired once their rubric pass rates reach a threshold:
\begin{equation}
    \mathcal{P}_k^{\mathrm{retire}}
    = \left\{
        (r_i,s_i)\in\mathcal{P}_k
        \mid \hat p_i\geq\eta_{\mathrm{high}}
      \right\},
    \qquad \Delta_{\mathrm{pool}}\mid(k+1),
    \label{eq:pair_retirement}
\end{equation}
where $\eta_{\mathrm{high}}\in(0,1]$ is the retirement threshold.
The paired pool evolves as
\begin{equation}
    \mathcal{P}_{k+1}
    = \bigl(\mathcal{P}_k\setminus\mathcal{P}_k^{\mathrm{retire}}\bigr)
    \cup\mathcal{P}_k^{\mathrm{new}},
    \qquad \Delta_{\mathrm{pool}}\mid(k+1).
    \label{eq:pool_update}
\end{equation}
Between these scheduled updates, the pool membership remains fixed.

\paragraph{Skill Activation and Refinement.}
Persistently low rubric pass rates motivate additional guidance for exploration.
At scheduled updates, paired skills are activated to guide alternative actions and hidden as performance improves:
\begin{equation}
    \begin{aligned}
        \mathcal{S}_k^{\mathrm{activate}}
        &= \{s_i\in\mathcal{S}_k^{\mathrm{hidden}}
            \mid \hat p_i\leq\eta_{\mathrm{low}}\},\\
        \mathcal{S}_k^{\mathrm{hide}}
        &= \{s_i\in\mathcal{S}_k^{\mathrm{active}}
            \mid \eta_{\mathrm{low}}<\hat p_i<\eta_{\mathrm{high}}\},
    \end{aligned}
    \qquad \Delta_{\mathrm{pool}}\mid(k+1),
    \label{eq:skill_activation}
\end{equation}
where $\mathcal{S}_k^{\mathrm{hidden}}$ and $\mathcal{S}_k^{\mathrm{active}}$ denote the currently hidden and active skills, respectively, and $\eta_{\mathrm{low}}\in[0,\eta_{\mathrm{high}})$ is the activation threshold.
Hiding a skill simply deactivates its guidance while retaining its pair in the pool.
If failures persist despite active guidance ($\hat p_i\leq\eta_{\mathrm{low}}$), the reflection model refines the skill using recent failure evidence $\mathcal{B}_k$ and the existing pair:
\begin{equation}
    s_i'
    = G_{\mathrm{ref}}\!\left(\mathcal{B}_k,r_i,s_i\right),
    \qquad \Delta_{\mathrm{pool}}\mid(k+1).
    \label{eq:skill_refinement}
\end{equation}
The revised skill $s_i'$ replaces $s_i$ in the pair without changing its rubric.
After these updates, only the currently active skills are injected into the system prompt for subsequent rollouts.

\subsubsection{Capability-Based Adaptive Sampling}
Guidance changes how the policy explores, but improvement also requires tasks that exercise the relevant behaviors.
Capability-based adaptive sampling therefore uses rubric evaluations to allocate training experience toward task types where specific capabilities remain weak.

\paragraph{Behavioral Evidence.}
Global capability estimates can obscure weaknesses specific to certain task types, whereas instance-level estimates fragment evidence across individual examples.
To strike a balance, we group tasks by their required operations or workflows, allowing related tasks to share behavioral evidence without collapsing differences across task types.
We assign these task-type tags prior to training using a predefined taxonomy that allows multiple tags per task (details are provided in Appendix~\ref{app:task_type_annotation}).
Each rubric is assigned to one of five predefined capabilities based on its target behavior (Appendix~\ref{app:adaptive_sampling}).
With task-type tags fixed, rubric evaluations update capability estimates within each task type $d$.
For capability $c$, the trajectory-level evidence from $\tau_j$ is calculated as:
\begin{equation}
    u_c(\tau_j)
    = \frac{1}{|\mathcal{R}_c(\tau_j)|}
      \sum_{r_i\in\mathcal{R}_c(\tau_j)} y_{ij},
    \label{eq:capability_evidence}
\end{equation}
where $\mathcal{R}_c(\tau_j)$ is the non-empty subset of active rubrics assigned to capability $c$ that apply to $\tau_j$, and $y_{ij}\in\{0,1\}$ denotes the verdict for rubric $r_i$ on that trajectory.
Each trajectory contributes success and failure evidence as follows:
\begin{equation}
    \Delta S_{d,c}=u_c(\tau_j),
    \qquad \Delta F_{d,c}=1-u_c(\tau_j).
    \label{eq:capability_evidence_contribution}
\end{equation}
These contributions are accumulated into $S_{d,c}$ and $F_{d,c}$ for each task-type tag $d$ associated with $\tau_j$, with older evidence discounted over the course of training.

\paragraph{Coverage-Guided Discovery.}
While task-type tags are available from the outset, behavioral evidence must be collected via rollouts.
At training iteration $k$, we measure evaluation completeness by averaging the fraction of evaluated active rubrics across tasks and capabilities:
\begin{equation}
    C_k
    = \frac{1}{|\mathcal{X}|\,|\mathcal{C}_k|}
      \sum_{x\in\mathcal{X}}\sum_{c\in\mathcal{C}_k}
      \frac{|\mathcal{E}_k(x)\cap\mathcal{R}_{k,c}|}
           {|\mathcal{R}_{k,c}|},
    \label{eq:rubric_evaluation_coverage}
\end{equation}
where $\mathcal{X}$ denotes the training task set, $\mathcal{C}_k$ the capability categories represented in the active pool, $\mathcal{R}_{k,c}\subseteq\mathcal{R}_k$ the active rubrics assigned to capability $c$, and $\mathcal{E}_k(x)$ the active rubrics already evaluated for task $x$.
If the pool is empty, we set $C_k=0$.
Based on this coverage, each batch position with available candidates takes the discovery path with probability:
\begin{equation}
    \rho_k=\max(\rho_{\min},1-C_k),
    \label{eq:capability_discovery_ratio}
\end{equation}
where $\rho_{\min}\in[0,1]$ is the minimum discovery probability.
Starting with pure discovery ($C_0=0$, $\rho_0=1$), this sampling favors less-evaluated tasks initially and gradually shifts toward adaptive selection as coverage increases.
Discovery sampling details are provided in Appendix~\ref{app:adaptive_sampling}.

\paragraph{Adaptive Task Selection.}
Both near-certain failure and near-certain success limit behavioral contrasts among rollouts.
This motivates prioritizing task-type--capability pairs with intermediate pass probabilities.
Following VADE~\citep{hu2025vade}, we represent uncertainty in the pass probability for each task-type--capability pair $(d,c)$ with a Beta distribution and draw an estimate $\tilde p_{d,c}$:
\begin{equation}
    \tilde p_{d,c}\sim \operatorname{Beta}(1+S_{d,c},1+F_{d,c}),
    \label{eq:capability_sampling_priority}
\end{equation}
where the unit offsets reflect a uniform $\operatorname{Beta}(1,1)$ prior.
At iteration $k$, $\mathcal{A}_k$ denotes the set of eligible task-type--capability pairs.
For each $(d,c)\in\mathcal{A}_k$, the adaptive selection probability is defined as:
\begin{equation}
    q_k^{\mathrm{pair}}(d,c)
    = \frac{w_{d,c}}{\sum_{(d',c')\in\mathcal{A}_k}w_{d',c'}},
    \qquad w_{d,c}=\tilde p_{d,c}(1-\tilde p_{d,c})^2,
    \label{eq:capability_pair_selection}
\end{equation}
where $w_{d,c}$ is the unnormalized sampling weight.
A task is then sampled uniformly from the available members of the selected pair.
As rubrics update, we preserve evidence for unchanged criteria, discard obsolete contributions, and introduce fresh discovery needs for new criteria.
Evidence decay and batch construction details are provided in Appendix~\ref{app:adaptive_sampling}.

\subsection{Training Strategy}
Our training strategy combines task outcomes and rubric feedback through separately normalized advantages to guide policy optimization.

\paragraph{Advantage Estimation.}
Trajectories with identical task outcomes can differ substantially in behavioral quality.
To incorporate these differences without assuming a common reward scale, we normalize task rewards and rubric verdicts separately within each same-task rollout group:
\begin{equation}
    A_j^{\mathrm{task}}
    = \frac{R_j-\mu_R}{\sigma_R+\epsilon},
    \qquad
    A_{ij}^{\mathrm{rub}}
    = \frac{y_{ij}-\mu_i}{\sigma_i+\epsilon},
    \label{eq:separate_advantage}
\end{equation}
where $R_j$ is the length-regularized task reward, with the length penalty detailed in Appendix~\ref{app:policy_optimization}, and $\epsilon>0$ ensures numerical stability.
The statistics $(\mu_R,\sigma_R)$ are computed over the entire group, whereas $(\mu_i,\sigma_i)$ are computed only over trajectories with applicable evaluations for rubric $r_i$.
Let $\mathcal{I}$ denote the set of rubrics exhibiting non-zero verdict variance within the group.
For trajectories with usable feedback from $\mathcal{I}$, the combined advantage is:
\begin{equation}
    A_j=(1-\lambda)A_j^{\mathrm{task}}
    +\frac{\lambda}{|\mathcal{I}|}
      \sum_{i\in\mathcal{I}} A_{ij}^{\mathrm{rub}},
    \label{eq:combined_advantage}
\end{equation}
where $\lambda\in[0,1]$ controls the contribution of behavioral feedback.
Inapplicable evaluations are excluded from normalization and contribute zero rubric advantage.
If a trajectory lacks applicable feedback from $\mathcal{I}$, it simply relies on $A_j=A_j^{\mathrm{task}}$.

\paragraph{Policy Optimization.}
The combined advantage unifies task completion and behavioral quality into a single policy update. 
We use GRPO-style policy optimization~\citep{shao2024deepseekmath} with importance-ratio filtering, applying the trajectory advantage exclusively to model-generated tokens while excluding tool outputs and environment observations (details are provided in Appendix~\ref{app:policy_optimization}).

\begin{table}[t]
    \caption{Pass rates (\%) on SkillsBench v1.1 and Terminal-Bench v2.1 (TB).
    SkillsBench categories are software engineering (SE), natural science (NS), office and white collar (OW), industrial and physical systems (IP), finance and economics (FE), mathematics and operations research/formal reasoning (MR), cybersecurity (CS), and media and content production (MC).
    For SkillsBench, GPT-5.5 results are from the official leaderboard; dashes denote models without reported results from either that leaderboard or our own evaluation.
    \textbf{Bold values} indicate the best results within comparable-scale models.
    }
    \label{tab:main_results}
    \begin{center}
        \footnotesize
        \setlength{\tabcolsep}{5pt}
        \resizebox{\linewidth}{!}{%
            \begin{tabular}{l*{10}{c}}
                \toprule
                & \multicolumn{9}{c}{\textbf{SkillsBench v1.1}} & \textbf{TB v2.1} \\
                \cmidrule(lr){2-10} \cmidrule(l){11-11}
                \textbf{Model} & SE & NS & OW & IP & FE & MR & CS & MC & \textbf{Overall} & \textbf{Overall} \\
                \midrule
                \multicolumn{11}{l}{\textcolor{gray}{\textit{Proprietary Models}}} \\
                GPT-5.4 Mini & 27.1 & 35.7 & 45.2 & 21.4 & 25.9 & 41.7 & 33.3 & 66.7 & 34.5 & 59.2 \\
                GPT-5.5 & 63.4 & 77.9 & 76.2 & 57.5 & 37.0 & 95.0 & 69.0 & 60.0 & 67.3 & 84.3 \\
                Claude Opus-4.7 & 58.3 & 83.3 & 54.8 & 54.8 & 44.4 & 50.0 & 57.1 & 53.3 & 58.6 & 83.1 \\
                \midrule
                \multicolumn{11}{l}{\textcolor{gray}{\textit{Open-Weight Models}}} \\
                GPT-OSS-120B & -- & -- & -- & -- & -- & -- & -- & -- & -- & 26.2 \\
                MiniMax-M2.7 & 14.6 & 59.5 & 28.6 & 19.0 & 33.3 & 29.2 & 23.8 & 13.3 & 28.7 & 55.4 \\
                GLM-5.1 & 41.7 & 76.2 & 69.0 & 45.2 & 33.3 & 37.5 & 47.6 & 80.0 & 53.6 & 61.8 \\
                Kimi-K2.6 & 39.6 & 76.2 & 57.1 & 50.0 & 44.4 & 66.7 & 47.6 & 66.7 & 55.2 & 65.9 \\
                DeepSeek-V3.2 & -- & -- & -- & -- & -- & -- & -- & -- & -- & 46.8 \\
                DeepSeek-V4-Pro & 37.5 & 73.8 & 52.4 & 42.9 & 29.6 & 66.7 & 42.9 & 80.0 & 51.3 & 64.8 \\
                DeepSeek-V4-Pro-0813 & 62.5 & 81.0 & 64.3 & 47.6 & 48.1 & 50.0 & 42.9 & 60.0 & 59.0 & 78.7 \\
                Qwen3.5-122B-A10B & 18.8 & 40.5 & 23.8 & 16.7 & 22.2 & 33.3 & 19.0 & 13.3 & 24.1 & 47.6 \\
                Qwen3.5-397B-A17B & 16.7 & 45.2 & 38.1 & 31.0 & 22.2 & 37.5 & 23.8 & 20.0 & 30.3 & 51.3 \\
                Nemotron-3-Ultra-550B-A55B & -- & -- & -- & -- & -- & -- & -- & -- & -- & 53.9 \\
                \midrule
                \multicolumn{11}{l}{\textcolor{gray}{\textit{Comparable-Scale Models}}} \\
                Qwen3.5-27B (Base Model) & 14.6 & 40.5 & 26.2 & 23.8 & 18.5 & 29.2 & 14.3 & 6.7 & 23.4 & 41.6 \\
                VADE & 25.0 & 54.8 & 38.1 & 31.0 & 33.3 & \textbf{50.0} & 23.8 & 73.3 & 38.7 & 43.8 \\
                OnlineRubrics & 29.2 & \textbf{64.3} & 38.1 & 31.0 & 37.0 & 33.3 & \textbf{28.6} & 73.3 & 40.2 & 46.1 \\
                RuscaRL & 33.3 & 57.1 & 42.9 & 26.2 & \textbf{40.7} & 29.2 & 23.8 & 73.3 & 39.5 & 44.9 \\
                \rowcolor{red!10}
                \textbf{\method} & \textbf{41.7} & 59.5 & \textbf{52.4} & \textbf{38.1} & \textbf{40.7} & 33.3 & 23.8 & \textbf{80.0} & \textbf{45.6} & \textbf{50.6} \\
                \bottomrule
            \end{tabular}%
        }
    \end{center}
\end{table}

\section{Experiment}
In this section, we evaluate \method by addressing three key research questions:
\textbf{RQ1:} Can \method improve agentic task performance via rubric--skill co-evolution and capability-based adaptive sampling? (Section~\ref{sec:main_results})
\textbf{RQ2:} Does rubric--skill co-evolution sustain informative behavioral feedback and enhance exploration? (Section~\ref{sec:model_analysis})
\textbf{RQ3:} How does capability-based adaptive sampling allocate training data, and does \method improve training efficiency over outcome-only RL? (Section~\ref{sec:model_analysis})

\subsection{Experiment Settings}
\minisection{Training Setup}
We train on 1,728 tasks from our constructed corpus (Appendix~\ref{app:training_corpus}), with model configurations and hyperparameters detailed in Appendix~\ref{app:model_configurations}.

\minisection{Benchmarks}
We evaluate on two challenging long-horizon agent benchmarks: SkillsBench v1.1~\citep{li2026skillsbench}, covering expertise-intensive workflows, and Terminal-Bench v2.1~\citep{merrill2026terminal}, covering command-line tasks.
Crucially, the skills evolved by \method during training are withheld during evaluation.
Further benchmark and protocol details are provided in Appendix~\ref{app:evaluation_details}.

\minisection{Baselines}
We include proprietary models GPT-5.4 Mini, GPT-5.5, and Claude Opus-4.7, alongside larger open-weight models GPT-OSS-120B~\citep{openai2025gptoss120bgptoss20bmodel}, MiniMax-M2.7~\citep{chen2026minimax}, GLM-5.1~\citep{zeng2026glm}, Kimi-K2.6~\citep{team2025kimi}, DeepSeek-V3.2~\citep{liu2025deepseek}, DeepSeek-V4-Pro~\citep{xu2026deepseek}, Qwen3.5-122B-A10B~\citep{qwen35blog} and Qwen3.5-397B-A17B, and Nemotron-3-Ultra-550B-A55B~\citep{blakeman2026nemotron} as performance references.
Comparable-scale baselines include our base model, Qwen3.5-27B~\citep{qwen35blog}, as well as VADE~\citep{hu2025vade}, OnlineRubrics~\citep{rezaei2025online}, and RuscaRL~\citep{zhou2025ruscarl}.
Baseline configurations are in Appendix~\ref{app:baseline_configurations}.

\minisection{Metrics}
We report task pass rates (\%) evaluated by the official benchmark verifiers.
Results for SkillsBench include both domain-level and overall pass rates, whereas Terminal-Bench performance is summarized by a single overall rate.
For our own evaluations, pass rates are averaged over three independent runs (aggregation details in Appendix~\ref{app:eval_metrics}).

\subsection{Main Results}
\label{sec:main_results}

\textit{\method outperforms all comparable-scale baselines on both benchmarks, achieves competitive performance with substantially larger models, and transfers effectively to Terminal-Bench under a different agent harness.}

Table~\ref{tab:main_results} shows that \method achieves the highest overall pass rates among comparable-scale models on both benchmarks.
On SkillsBench, \method improves over the base model across all evaluated domains and achieves leading results in most domains among comparable-scale models.
\method also surpasses the larger Qwen3.5 variants on SkillsBench while remaining competitive with them on Terminal-Bench, highlighting the scope for agentic training to narrow the performance gap with substantially larger models.
This advantage is consistent across the three SkillsBench evaluation runs (Appendix~\ref{app:stability}).

Beyond generalization to unseen tasks, \method also transfers across agent harnesses: training uses Kilo Code, whereas evaluation on Terminal-Bench uses Terminus-2.
Under this change, \method maintains its advantage over the base model and all comparable-scale baselines, indicating that the training gains extend beyond a single benchmark and execution framework.

\subsection{Model Analysis}
\label{sec:model_analysis}

\begin{figure}[htbp]
    \centering
    \includegraphics[width=\linewidth]{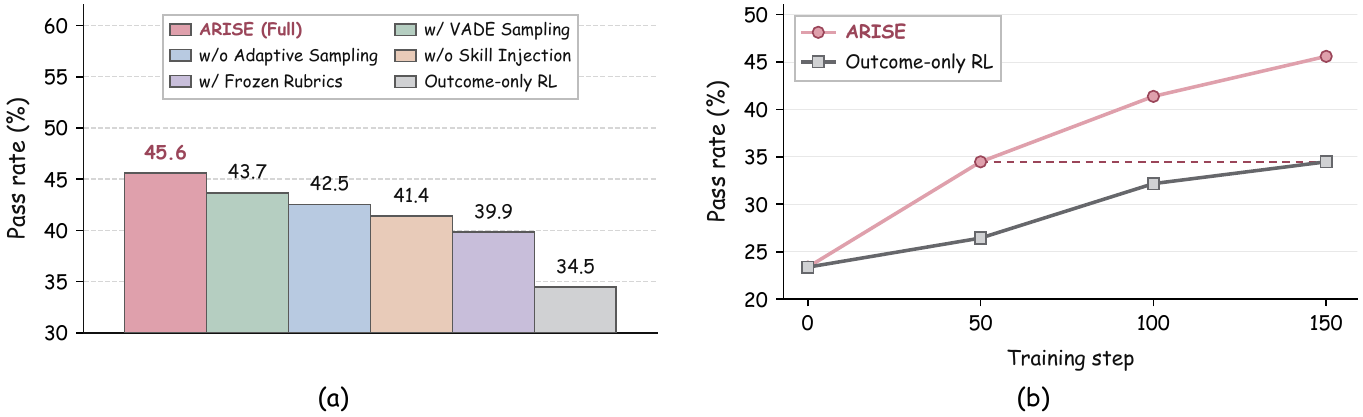}
    \vspace{-10pt}
    \caption{Ablation results and training efficiency on SkillsBench v1.1. (a) Pass rates for \method and its ablated variants. (b) Pass rates over training under the same per-step rollout budget.}
    \label{fig:ablation_efficiency}
\end{figure}

\minisection{Ablation Study}
\textit{Dynamic rubric evolution, skill-guided exploration, and capability-based sampling each contribute to the performance of \method.}
We evaluate five variants with matched initialization, data, and training budgets:
(i) \textit{Frozen Rubrics}: fixes the pool to include all rubrics from \method's entire training lifecycle;
(ii) \textit{w/o Skill Injection}: removes skill guidance from the policy context;
(iii) \textit{w/o Adaptive Sampling}: disables our adaptive sampler;
(iv) \textit{w/ VADE Sampling}: replaces our sampler with the outcome-based VADE approach~\citep{hu2025vade};
(v) \textit{Outcome-only RL}: removes rubric rewards, skill injection, and adaptive sampling.
As shown in Figure~\ref{fig:ablation_efficiency}(a), \method outperforms all variants.
Notably, even access to the full lifecycle rubric pool cannot replace dynamic updates, and removing skill guidance degrades performance despite retaining rubric rewards.
Furthermore, while VADE sampling improves upon non-adaptive sampling, it falls short of our capability-based approach, confirming the value of behavioral evidence beyond mere task outcomes.

\begin{figure}[htbp]
    \centering
    \includegraphics[width=\linewidth]{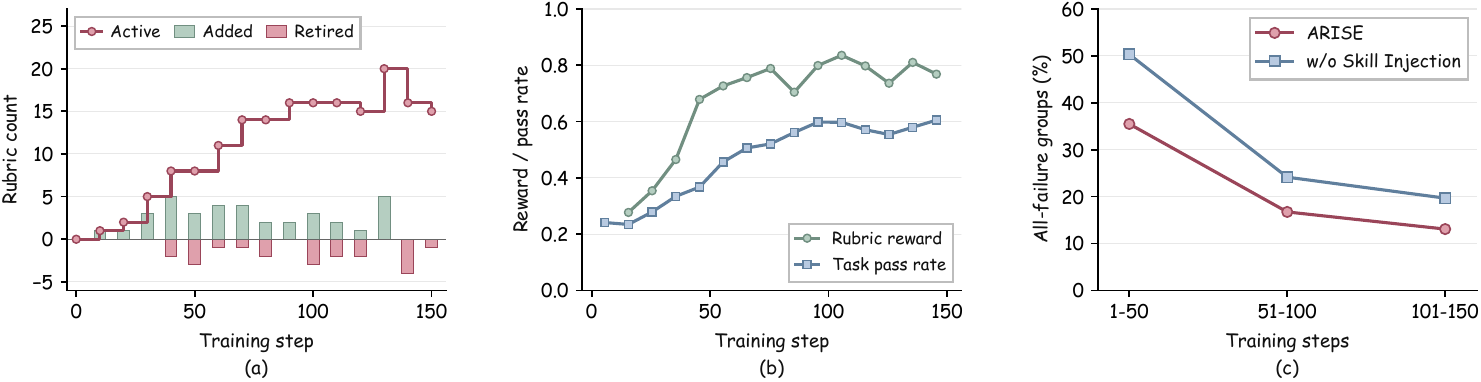}
    \vspace{-10pt}
    \caption{Behavioral feedback and exploration during training. (a) Active, added, and retired rubric counts. (b) Mean rubric reward and verifier-based task pass rate in 10-step windows. (c) All-failure rollout group rates, averaged equally across tasks in the same fixed task set.}
    \label{fig:coevo_dynamics}
\end{figure}

\minisection{Behavioral Feedback and Exploration}
\textit{\method continually refreshes its behavioral criteria and exhibits fewer all-failure rollout groups than training without skill injection.}
Starting from an empty pool, rubric evolution introduces 36 criteria and retires 21 during training, leaving 15 active (Figure~\ref{fig:coevo_dynamics}(a)).
This turnover demonstrates that supervision evolves through dynamic replacement rather than mere accumulation.
Figure~\ref{fig:coevo_dynamics}(b) shows an early rise in mean rubric reward followed by fluctuations under the changing rubric pool, alongside an overall increase in verifier-based task pass rates.
Notably, behavioral pass rates remain high after rubric retirement, even without skill guidance (Appendix~\ref{app:retention}).
To evaluate exploration, we compare \method against the no-skill variant on a fixed task set.
\method consistently yields lower all-failure rollout rates across all training stages, with the most pronounced gap appearing early on (Figure~\ref{fig:coevo_dynamics}(c)).
Together with the skill-injection ablation, this confirms that skill guidance crucially helps the policy discover successful trajectories.
Appendix~\ref{app:evolution_cases} provides concrete examples of individual rubric--skill lifecycles.

\begin{figure}[htbp]
    \centering
    \includegraphics[width=\linewidth]{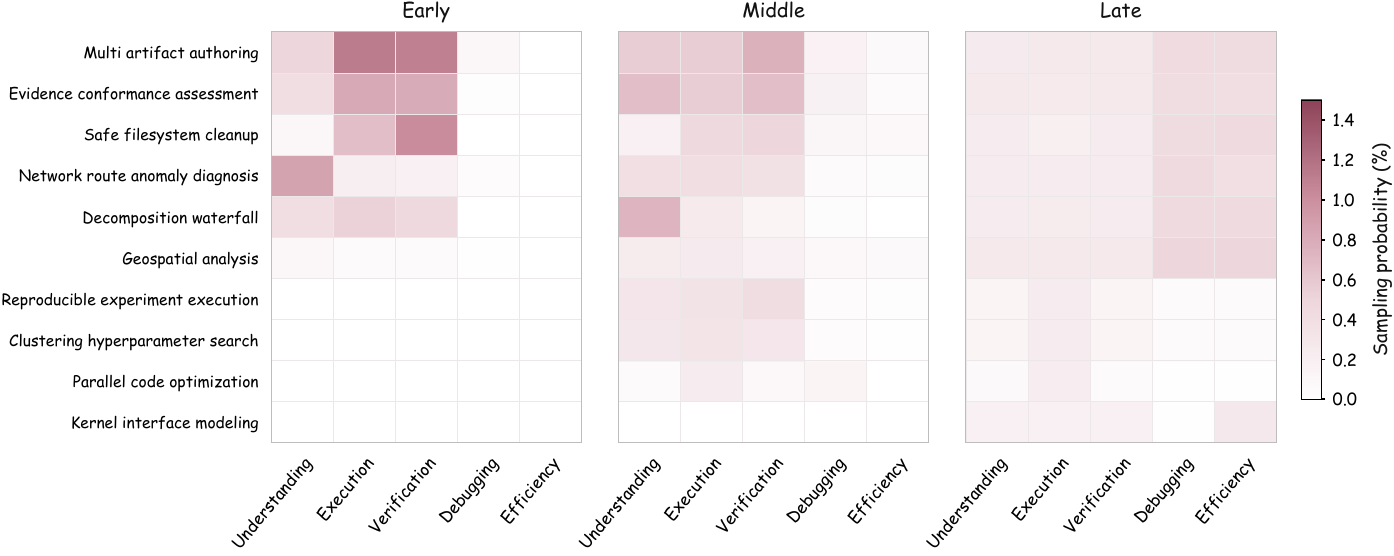}
    \vspace{-10pt}
    \caption{Adaptive sampling probabilities for selected task types and behavioral capabilities, excluding discovery sampling. Early, middle, and late stages correspond to training steps 1--50, 51--100, and 101--150, respectively.}
    \label{fig:task_heatmap}
\end{figure}

\minisection{Training Data Allocation and Efficiency}
\textit{Behavioral evidence induces stage-dependent sampling priorities, while \method reaches comparable task performance with fewer training steps than outcome-only RL.}
Figure~\ref{fig:task_heatmap} illustrates adaptive sampling probabilities for selected task types and capabilities across early, middle, and late training stages (excluding discovery sampling).
Blank cells in early training reflect the initially empty pool and limited evaluation coverage, as many task-type--capability pairs remain unexplored or lack active rubrics. 
However, tasks associated with these blank cells are still sampled via discovery.
Early sampling emphasizes execution and verification, whereas later stages prioritize debugging and efficiency as new rubrics for these capabilities enter the active pool.
Coupled with rising task pass rates, this shift indicates that once the policy achieves basic execution competence, training increasingly targets error correction and workflow efficiency.
At the framework level, Figure~\ref{fig:ablation_efficiency}(b) demonstrates \method's superior sample efficiency: under the same per-step rollout budget, it achieves the step-150 pass rate of outcome-only RL by step 50 and continues improving thereafter.
This represents a substantial reduction in required training steps and rollout budget (see Appendix~\ref{app:training_time} for training time analysis).

\section{Conclusion}
We introduced \method, an agentic reinforcement learning framework that adapts the learning process to evolving capability gaps.
By turning rollout evidence into evolving rubrics and paired skills, it connects the identification of behavioral weaknesses with targeted exploration and task selection.
Experiments on SkillsBench and Terminal-Bench show improvements over all evaluated comparable-scale baselines, competitive performance with substantially larger models, and effective transfer across agent harnesses.
The analysis further demonstrates improved training efficiency under a matched per-step rollout budget.
Together, these findings highlight the value of continually adapting both behavioral feedback and learning opportunities as agent capabilities develop.

\section*{Acknowledgments}
The research was supported by the National Natural Science Foundation of China (Grant No. 62406193) and the ShanghaiTech AI Initiative (Grant No. AI2026B08).
The authors also gratefully acknowledge assistance from the Key Laboratory of Intelligent Perception and Human-Machine Collaboration (ShanghaiTech University), Ministry of Education, and the HPC Platform of ShanghaiTech University.
This work was also supported by Ant Group Research Intern Program.

\bibliography{main}
\bibliographystyle{main}

\clearpage

\appendix
\section{Limitations}
\label{app:limitations}

In this work, we train exclusively on Qwen3.5-27B, leaving the effectiveness of \method across other model scales and architectures unverified.
Furthermore, evaluation is limited to SkillsBench and Terminal-Bench, deferring generalization to diverse agent environments and real-world workflows to future work.
Beyond these empirical limitations, rubric generation and evaluation rely heavily on the reflection and judge models.
Consequently, any inherent errors or biases in these models may compromise the resulting behavioral supervision and training decisions.

\section{Implementation Details}
\label{app:implementation}
\subsection{Training Procedure}
\label{app:training_procedure}

Algorithm~\ref{alg:arise_training} summarizes how \method starts from an empty pool and uses initial rollout evidence to generate rubric--skill pairs.
As rubric evaluations accumulate, capability-based sampling and selectively activated skills guide further rollouts, whose task rewards and rubric judgments support policy updates.
Periodic pool updates add new rubric--skill pairs, refine skill guidance, and retire mastered criteria as training progresses.

\begin{algorithm}[htbp]
\caption{\method Training Procedure}
\label{alg:arise_training}
\begin{algorithmic}[1]
\REQUIRE Training tasks $\mathcal{X}$ with fixed task-type tags; initial policy $\pi_\theta$; judge model; reflection model $G_{\mathrm{ref}}$; rollout group size $G$; configuration in Table~\ref{tab:training_configuration}
\ENSURE Trained policy $\pi_\theta$
\STATE Initialize the pool of rubric--skill pairs $(r_i,s_i)$: $\mathcal{P}_0\leftarrow\emptyset$
\STATE Initialize success evidence $S_{d,c}\leftarrow 0$ and failure evidence $F_{d,c}\leftarrow 0$ for each task type $d$ and capability $c$, and evaluation coverage $C_0\leftarrow 0$
\FOR{each training iteration $k=0,1,\ldots$ until the training budget is exhausted}
    \STATE Set discovery probability $\rho_k\leftarrow\max(\rho_{\min},1-C_k)$
    \STATE Sample $B$ tasks by mixing discovery and adaptive selection with $\rho_k$ (Appendix~\ref{app:adaptive_sampling})
    \STATE Generate $G$ rollouts per task, injecting currently active skills into the system prompt
    \STATE Obtain verifier rewards $R_{\mathrm{task}}(x,\tau_j)$ and record interaction evidence
    \IF{$\mathcal{P}_k=\emptyset$}
        \STATE Extract initial behavioral diagnostics without rubric verdicts
    \ELSE
        \STATE Judge rubric applicability and verdicts $y_{ij}$, and identify uncovered behaviors
    \ENDIF
    \STATE Collect recent diagnostics and contextual information in $\mathcal{B}_k$
    \STATE Apply length regularization and compute $A_j$ using Equation~\ref{eq:combined_advantage}, with task-only fallback when needed
    \STATE Update $\theta$ using the policy objective in Appendix~\ref{app:policy_optimization}
    \STATE Update rubric-level evidence and aggregate $S_{d,c}$ and $F_{d,c}$ (Appendix~\ref{app:adaptive_sampling}); refresh coverage $C_{k+1}$
    \STATE Set $\mathcal{P}_{k+1}\leftarrow\mathcal{P}_k$
    \IF{$(k+1)\bmod\Delta_{\mathrm{pool}}=0$}
        \STATE Estimate pass rates $\hat p_i$ for rubrics $r_i$ with applicable evaluations
        \STATE At $\hat p_i\leq\eta_{\mathrm{low}}$, activate hidden skills or replace already active skills with refined versions within their pairs
        \STATE At $\eta_{\mathrm{low}}<\hat p_i<\eta_{\mathrm{high}}$, hide active skills while retaining their pairs
        \STATE At $\hat p_i\geq\eta_{\mathrm{high}}$, remove the corresponding pairs from $\mathcal{P}_{k+1}$
        \STATE Generate $\mathcal{P}_k^{\mathrm{new}}$ with $G_{\mathrm{ref}}$ from $\mathcal{B}_k$ and $\mathcal{P}_k$, subject to pool limits
        \STATE Assign each new rubric to a predefined capability category based on its target behavior
        \STATE Set $\mathcal{P}_{k+1}\leftarrow\mathcal{P}_{k+1}\cup\mathcal{P}_k^{\mathrm{new}}$, with new skills initially hidden
        \STATE Preserve unchanged rubric evidence, discard obsolete contributions, and refresh coverage $C_{k+1}$
    \ENDIF
\ENDFOR
\RETURN $\pi_\theta$
\end{algorithmic}
\end{algorithm}

\subsection{Model Configurations}
\label{app:model_configurations}

We initialize \method with Qwen3.5-27B and train it using AReaL 2.0~\citep{fu2026areal}, an asynchronous reinforcement learning framework with decoupled services for agent execution, inference, and policy optimization.
DeepSeek-V4-Flash~\citep{xu2026deepseek} serves as both the judge model for rubric evaluation and the reflection model for rubric--skill generation and refinement, with its parameters held fixed throughout training.
Training runs on 32 NVIDIA H800 GPUs for approximately two days.
Table~\ref{tab:training_configuration} summarizes the detailed training and co-evolution configurations.

\begin{table}[htbp]
    \caption{Training and co-evolution configuration.}
    \label{tab:training_configuration}
    \begin{center}
        \begin{tabular}{@{}cc@{}}
            \toprule
            \textbf{Configuration} & \textbf{Value} \\
            \midrule
            Training steps & 150 \\
            Tasks per step $B$ & 32 \\
            Rollouts per task $G$ & 8 \\
            Context length & 131,072 tokens \\
            Sampling temperature & 1.0 \\
            Top-$p$ & 1.0 \\
            Learning rate & $2\times10^{-6}$ \\
            Warmup steps & 40 \\
            Learning-rate schedule & Constant after warmup \\
            Precision & BF16 \\
            Rubric advantage weight $\lambda$ & 0.3 \\
            Pool update interval $\Delta_{\mathrm{pool}}$ & 10 training steps \\
            Active pool size $|\mathcal{P}_k|$ & $\leq 64$ \\
            New pairs per update $|\mathcal{P}_k^{\mathrm{new}}|$ & $\leq 8$ \\
            Activation/refinement threshold $\eta_{\mathrm{low}}$ & 0.3 \\
            Retirement threshold $\eta_{\mathrm{high}}$ & 0.9 \\
            \bottomrule
        \end{tabular}
    \end{center}
\end{table}

\subsection{Rollout Evidence}
\label{app:rollout_evidence}

For each rollout, we retain the interaction trace, including task context, agent responses, tool calls and their arguments, tool outputs, and environment feedback, together with verifier outcomes and information about available skills.
Trajectory analysis and rubric evaluation use the interaction trace as their primary evidence, with final task outcomes providing auxiliary context rather than determining behavioral judgments.
The resulting diagnostics identify observable behavioral gaps and useful strategies, their supporting signals, and whether existing rubrics already cover them.

Pool updates aggregate evidence over the preceding ten training steps.
Each generation call selects up to 256 records containing diagnostic evidence, without stratifying by task outcome.
The reflection model receives these compressed diagnostics, skill-use summaries, and task and pool metadata rather than full trajectories or per-rollout task rewards and outcome labels.
Issue and contrast evidence support the failure and success conditions of each new rubric, respectively.
Skill refinement additionally receives the previous skill and its paired criterion and rubric, using failed-task evidence and outcome context to improve guidance when the targeted behavior remains unresolved.
Prompt templates are provided in Appendix~\ref{app:prompts}.

\subsection{Policy Optimization}
\label{app:policy_optimization}

The policy update uses the combined trajectory advantage $A_j$ from Equation~\ref{eq:combined_advantage}, with the task-only fallback described in the main text.
We sample eight trajectories per task and use group-relative reward normalization without a learned value function.

\paragraph{Length Regularization.}
To control generation length during RL, as also explored in Kimi K2~\citep{team2025kimi}, we adjust positive task rewards using relative generation lengths within each task group.
Let $R_j^{\mathrm{raw}}=R_{\mathrm{task}}(x,\tau_j)$ and let $\ell_j$ count model-generated tokens in $\tau_j$, excluding the task prompt, tool outputs, and environment observations.
Regularization is active only when the group contains both positive and non-positive task rewards, its maximum generation length exceeds a threshold $L_{\mathrm{th}}$, and its generation lengths are not all equal.
The task reward used in Equation~\ref{eq:separate_advantage} is
\begin{equation}
    R_j =
    \begin{cases}
        R_j^{\mathrm{raw}}\!\left[1+\alpha\!\left(
        1-2\dfrac{\ell_j-\ell_{\min}}
        {\ell_{\max}-\ell_{\min}+\epsilon}\right)\right],
        & \text{if active and }R_j^{\mathrm{raw}}>0,\\
        R_j^{\mathrm{raw}}, & \text{otherwise},
    \end{cases}
    \label{eq:length_regularized_reward}
\end{equation}
where $\ell_{\min}$ and $\ell_{\max}$ are the minimum and maximum generation lengths over the entire task group, including unsuccessful trajectories, and $\alpha$ controls the adjustment strength.
We set $L_{\mathrm{th}}=16{,}384$ tokens and $\alpha=0.5$, with $\epsilon=10^{-9}$ ensuring numerical stability both here and in the advantage normalization of Equation~\ref{eq:separate_advantage}.
Thus, shorter successful trajectories receive larger relative rewards, while non-positive rewards remain unchanged.
This adjustment precedes group normalization and does not modify rubric verdicts.

\paragraph{Optimization Objective.}
Asynchronous rollouts may be generated by an earlier policy, so the update accounts for the difference between the current policy and the policy that sampled each token.
We use $u$ to index tokens, distinct from the interaction-step index $t$ in the problem formulation.
For a model-generated token $z_{j,u}$ and its preceding interaction context $h_{j,u}$, define the importance ratio
\begin{equation}
    \omega_{j,u}(\theta)
    = \frac{\pi_\theta(z_{j,u}\mid h_{j,u})}
           {\pi_{\mathrm{beh},j,u}(z_{j,u}\mid h_{j,u})},
    \label{eq:policy_importance_ratio}
\end{equation}
where the denominator is the token likelihood recorded during rollout generation, and $\pi_{\mathrm{beh},j,u}$ denotes the corresponding behavior policy.
Let $m_{j,u}$ indicate a model-generated token and let $\widetilde m_{j,u}=m_{j,u}\mathbf{1}\{0.5<\omega_{j,u}(\theta)<5.0\}$ retain tokens within the configured importance-ratio bounds.
The mask, rollout likelihoods, and trajectory advantages are held fixed during differentiation.
With the configured mask-based update, the minimized loss is equivalent to
\begin{equation}
    \mathcal{L}_{\mathrm{policy}}(\theta)
    = \frac{1}{\sum_{j,u}m_{j,u}}
      \sum_{j,u}\widetilde m_{j,u}
      \min\!\left\{
          -\omega_{j,u}(\theta)A_j,\;\kappa|A_j|
      \right\},
    \label{eq:policy_optimization_loss}
\end{equation}
where the sums span the training batch and $\kappa=3$ is the dual-clipping coefficient.
Dual clipping caps the loss for negative advantages, while leaving the importance-weighted term unclipped for positive advantages.
Unlike the standard PPO surrogate, this implementation rejects out-of-range tokens rather than clipping their ratios, and it applies no reference-policy KL penalty.
Tool outputs and environment observations remain in the conditioning context but incur no direct loss.
The final loss is normalized by the total number of model-generated tokens before filtering and aggregated across all workers.
An update with no retained tokens yields a zero policy gradient.

\subsection{Adaptive Sampling}
\label{app:adaptive_sampling}

\paragraph{Capability Categories.}
Each rubric is assigned to one of the five categories in Table~\ref{tab:capability_categories} according to the primary behavior it evaluates, rather than the task topic or final outcome.
These categories define the capability index $c$ used in the main text.

\begin{table}[htbp]
    \caption{Behavioral capability categories used to group rubric evidence.}
    \label{tab:capability_categories}
    \begin{center}
        \resizebox{\linewidth}{!}{%
        \begin{tabular}{@{}>{\centering\arraybackslash}m{0.2\linewidth}>{\centering\arraybackslash}m{\dimexpr0.8\linewidth-2\tabcolsep\relax}@{}}
            \toprule
            \textbf{Capability} & \textbf{Behavioral focus} \\
            \midrule
            Understanding & Interpreting task requirements, constraints, and available information. \\
            \addlinespace
            Execution & Carrying out intended actions through appropriate tool use and artifact manipulation. \\
            \addlinespace
            Verification & Checking results against requirements and grounding completion claims in observable evidence. \\
            \addlinespace
            Debugging & Diagnosing failures, identifying their causes, and applying targeted corrections. \\
            \addlinespace
            Efficiency & Avoiding redundant work and unnecessary resource use while preserving task progress. \\
            \bottomrule
        \end{tabular}%
        }
    \end{center}
\end{table}

\paragraph{Evidence Maintenance.}
To invalidate evidence selectively when rubrics change, we retain separate success and failure accumulators $S_{d,i}$ and $F_{d,i}$ for each task-type tag $d$ and rubric $r_i$.
The contribution in Equation~\ref{eq:capability_evidence_contribution} is distributed across the applicable rubric evaluations used to compute it: each verdict contributes $y_{ij}/|\mathcal{R}_c(\tau_j)|$ and $(1-y_{ij})/|\mathcal{R}_c(\tau_j)|$ to its corresponding accumulators.
Contributions are summed over trajectories in each complete rollout group and applied to every task-type tag associated with the task.
Following the two-scale decay design of VADE~\citep{hu2025vade}, we update the stored accumulators when an observation arrives $\Delta v>0$ policy-version increments after the last evidence update:
\begin{equation}
    (S_{d,i},F_{d,i})
    \leftarrow
    \gamma_{\mathrm{obs}}\gamma_{\mathrm{unobs}}^{\Delta v-1}
    (S_{d,i},F_{d,i})
    +(\Delta S_{d,i},\Delta F_{d,i}),
    \label{eq:rubric_evidence_decay}
\end{equation}
where $\Delta S_{d,i}$ and $\Delta F_{d,i}$ are the newly received contributions, and $\Delta v$ measures elapsed policy versions rather than training iterations indexed by $k$.
We set $\gamma_{\mathrm{obs}}=0.2$ and $\gamma_{\mathrm{unobs}}=0.999$.
Additional observations at the same policy version are added without further decay.
Without a new observation, stored evidence contributes with a factor $\gamma_{\mathrm{unobs}}^{\Delta v}$ when queried.
Summing these current contributions over active rubrics assigned to capability $c$ yields $S_{d,c}$ and $F_{d,c}$; the unit Beta prior is not decayed.
Pool updates remove only the affected rubric evidence and coverage records, without renormalizing retained contributions.

\paragraph{Coverage and Discovery.}
Coverage in Equation~\ref{eq:rubric_evaluation_coverage} is measured against all active rubrics in $\mathcal{R}_{k,c}$, whereas $\mathcal{R}_c(\tau_j)\subseteq\mathcal{R}_{k,c}$ contains only those applicable to trajectory $\tau_j$.
A rubric enters $\mathcal{E}_k(x)$ once all eight trajectories in a complete rollout group provide valid evaluations for it.
An explicit not-applicable verdict counts toward coverage but contributes no success or failure evidence.
The minimum discovery probability is $\rho_{\min}=0.1$.
Discovery samples eligible tasks with outstanding evaluations using weights proportional to $1/(1+a_x)$, where $a_x$ counts prior complete-group evaluation attempts regardless of judge success.

\paragraph{Batch Construction.}
For each batch position, the sampler mixes discovery and adaptive selection according to Equation~\ref{eq:capability_discovery_ratio} when discovery candidates are available.
The sampler redraws the Beta estimates for eligible pairs at each adaptive selection, then follows Equation~\ref{eq:capability_pair_selection} and samples uniformly from the available members of the selected pair.
Adaptive selection considers only tasks with prior applicable rubric evidence for the selected capability.
Tasks are selected without replacement within a batch.
Each pair may be targeted at most $\max(1,\lceil bB\rceil)$ times through adaptive selection, where $B$ is the task batch size and $b$ is the per-pair quota fraction.
We use $B=32$ and $b=0.25$, giving at most eight adaptive selections per target pair; discovery selections are not subject to this quota.
If adaptive selection is unavailable, the sampler uses inverse-attempt weights to select from remaining candidates, prioritizing discovery candidates.

\subsection{Prompt Templates}
\label{app:prompts}

We provide the prompts used for initial trajectory analysis, rubric--skill generation, rubric evaluation, skill refinement, and offline task-type annotation, with placeholders for input data and shared specifications.

\paragraph{Initial Trajectory Analysis.}
\label{app:prompt_initial_analysis}
Before the first rubric--skill pairs are generated, this prompt extracts diagnostic evidence from individual rollouts without assigning rubric verdicts.
The analysis identifies observable behavioral gaps and useful strategies.

\begin{tcolorbox}[
  breakable,
  boxrule=0.25pt,
  colback=promptBgSystem,
  colframe=black,
  colbacktitle=promptBgTitle,
  coltitle=black,
  title={Prompt: Initial Trajectory Analysis},
  fonttitle=\bfseries,
  fontupper=\small
]
\textbf{System prompt}

You analyze one trajectory to bootstrap a future rubric pool. There are no active rubrics yet, so do not produce verdicts. Return strict JSON with a required \texttt{diagnostics} object. These diagnostics may later be aggregated into future criteria and rubrics, but this prompt must only describe one completed rollout. Ground diagnostics primarily in \texttt{sample\_result.llm\_interaction\_trajectory}, which contains the exported raw LLM interaction trace for the rollout. Use \texttt{task\_skills} only as metadata about skills originally available in the task environment. Treat \texttt{verifier\_score}, \texttt{verifier\_success}, reward, final output, and \texttt{test\_stdout} as weak outcome context; they must not be the sole basis for a diagnostic item. Each item must describe one concrete, observable, reusable agent behavior pattern. Write item text in this shape: 'When \texttt{\textless{}condition\textgreater{}}, the agent should/failed to \texttt{\textless{}behavior\textgreater{}}, observable via \texttt{\textless{}signal\textgreater{}}.' Do not mention \texttt{verifier\_score}, reward, pass/fail, tests pass, or correct output as the reason. Do not infer hidden intent, understanding, carelessness, or motivation. Assign exactly one \texttt{issue\_tag} to every diagnostic item. Diagnostic item object contract: every item in \texttt{uncovered\_issues} and \texttt{positive\_uncovered\_strategies} must contain exactly these five fields: \texttt{kind}, \texttt{issue\_tag}, \texttt{text}, \texttt{observable\_signals}, and \texttt{related\_rubric\_ids}. \texttt{kind} must be \texttt{failure\_gap} for \texttt{uncovered\_issues} and \texttt{positive\_strategy} for \texttt{positive\_uncovered\_strategies}. \texttt{observable\_signals} and \texttt{related\_rubric\_ids} must be JSON arrays of strings. Prefer diagnostics that identify one reusable agent behavior pattern from the catalog below instead of restating exact task acceptance conditions.

\texttt{\{behavior\_issue\_taxonomy\}}

\texttt{\{behavior\_pattern\_catalog\}}

Third-party verifier constraint: final pytest/verifier results are post-hoc third-party evaluation signals. The agent cannot directly call hidden evaluator tests, hidden verifier internals, or a full hidden test suite unless a task-visible test command, script, verifier, or check is explicitly present in the task workspace. Use pytest/verifier outcomes only as evidence for task-visible behavior gaps. Criteria, rubrics, diagnostics, and skills must require only task-visible verification from visible files, commands, scripts, generated artifacts, explicit task constraints, or inspectable outputs.

Because this is bootstrap mode, \texttt{diagnostics.covered\_by\_active\_rubrics} must be \texttt{false} and every \texttt{related\_rubric\_ids} list must be empty. For any trajectory, \texttt{diagnostics.uncovered\_issues} may describe observable behavior gaps and \texttt{diagnostics.positive\_uncovered\_strategies} may describe useful behavior. Do not force either list when evidence is weak. Each list may contain at most 5 items. Each item text must be based only on observable \texttt{sample\_result} signals.

\textbf{User prompt}
\begin{verbatim}
{
  "sample_result": "{trajectory_result}",
  "output_schema": {
    "type": "object",
    "required": ["diagnostics"],
    "properties": {"diagnostics": "{diagnostics_schema}"}
  }
}
\end{verbatim}
\end{tcolorbox}

\paragraph{Rubric--Skill Generation.}
\label{app:prompt_pair_generation}
The reflection model receives compressed diagnostic evidence and descriptions of already covered criteria.
It generates paired rubrics and skills supported by contrasting behavioral evidence, without receiving per-trajectory task reward labels.

\begin{tcolorbox}[
  breakable,
  boxrule=0.25pt,
  colback=promptBgSystem,
  colframe=black,
  colbacktitle=promptBgTitle,
  coltitle=black,
  title={Prompt: Rubric--Skill Generation},
  fonttitle=\bfseries,
  fontupper=\small
]
\textbf{System prompt}

You are a trajectory reflection model. Your job is to compile compressed rollout evidence into a small global pool of paired criteria, rubrics, and skills. Return only strict JSON. Do not include markdown outside JSON, comments, rationale fields, or \texttt{rejection\_risk} fields.

\textbf{Generation contract:}

1. Produce at most \texttt{max\_items} items.

2. Each accepted item represents one global criteria and must derive exactly one rubric and one skill from that same criteria.

3. Each item must include \texttt{issue\_evidence\_refs}: evidence where the candidate issue is visibly present and a future rubric should fail. Each item must also include \texttt{contrast\_evidence\_refs} with at least one \texttt{evidence\_id} where the issue is visibly absent and a future rubric should pass.

4. Use only \texttt{evidence\_id} values present in the input. Never invent, rewrite, or summarize evidence refs.

5. \texttt{issue\_tag} must use exactly one tag from the trajectory behavior issue taxonomy below.

\texttt{\{behavior\_issue\_taxonomy\}}

5a. \texttt{capability\_tag} must classify the primary capability measured by \texttt{rubric\_rule\_text} using exactly one of: \texttt{\{capability\_categories\}}. Classify the rubric behavior itself, not the task category or final outcome.

6. \texttt{abstraction\_level} must be exactly \texttt{global}. Omit sample-specific, dataset-specific, task-family-only, or broad task-category observations.

7. Criteria must describe a concrete uncovered failure mode or positive strategy from \texttt{coevolution\_rubric\_diagnostics}, not a broad task category such as file extraction, data processing, or document formatting. Each criteria must be a one-sentence abstraction of a specific diagnostic item pattern; do not broaden a diagnostic about one behavior into a task category. Do not turn exact task requirements into criteria; abstract them into the underlying failure mechanism or reusable strategy. Avoid filename, required key, header string, quoted string, exact cell, slide, page, column, row, or variable-name literals in criteria and rubrics.

8. \texttt{observable\_signals} must include \texttt{coevolution\_rubric\_diagnostics} and may include \texttt{coevolution\_pool\_version}, \texttt{skill\_usage\_summary}, or \texttt{attempt\_group\_summary}. This field records generation provenance only; it does not define the runtime inputs or pass/fail predicate for the rubric judge. Do not rely on raw reasoning, raw tool arguments, full pytest output, final answers, or artifact text; those are audit-only fields and are intentionally absent here.

9. Per-evidence verifier, reward, success, and failure outcome labels are intentionally absent from this generation prompt. Do not infer hidden outcome labels, do not use final outcomes as the issue itself, and do not require criteria to separate passed and failed trajectories.

10. Each item must separate applicability from scoring. \texttt{applicability\_condition} must be one concrete, observable sentence describing the task scope or factual opportunity that makes the rubric relevant. It may reference an event such as an observed execution failure, but it must not depend on whether the agent performed the target behavior being scored. For a universally applicable rubric, write 'This rubric applies to every task.' Never use agent compliance, verifier outcome, reward, or hidden state as the applicability condition. A skipped required behavior must remain applicable and be judged by \texttt{rubric\_rule\_text}.

11. \texttt{rubric\_rule\_text} must be a binary pass/fail rule that a Reflection judge can apply directly to observable agent behavior in \texttt{sample\_result.llm\_interaction\_trajectory} and other task-visible \texttt{sample\_result} signals. Diagnostics are discovery provenance only, not judge inputs or conditions in the rubric rule. Never mention \texttt{coevolution\_rubric\_diagnostics}, \texttt{uncovered\_issues}, \texttt{positive\_uncovered\_strategies}, or another internal diagnostics field in \texttt{rubric\_rule\_text}. State the concrete action, omission, observation, tool result, or finalization behavior that makes the trajectory pass or fail. For example, do not write 'Fail if any \texttt{uncovered\_issue} describes repeated retries'; write 'Fail if the trajectory shows repeated retries without inspecting the concrete error between attempts; pass otherwise.' Do not include an inapplicable or \texttt{not\_applicable} branch in \texttt{rubric\_rule\_text}; that decision belongs exclusively to \texttt{applicability\_condition}. Never write rules that merely say the verifier passes, output is correct, tests pass, or \texttt{verifier\_score == 1.0}.

For any verification-type rubric, \texttt{rubric\_rule\_text} must state both (a) the observable trigger that makes verification necessary and (b) the sufficiency condition after which verification is complete. One appropriately scoped successful check after the latest relevant modification must be sufficient unless a new modification, failed check, source conflict, or anomaly creates new evidence. Never require unconditional repeated reading, recomputation, or the strongest possible verification for every artifact.

12. \texttt{skill\_text} must be an actionable Markdown skill for an agent to read from the skill pool. It must not describe reward math, hidden state, prompt injection, or a single sample solution. It must preserve the concrete behavior pattern from diagnostics.

13. Use \texttt{uncovered\_issues} and \texttt{positive\_uncovered\_strategies} from \texttt{coevolution\_rubric\_diagnostics} only as the discovery source for every criteria. Abstract their contents into direct trajectory behavior for \texttt{rubric\_rule\_text} instead of copying the diagnostics container or field names. Evidence without diagnostics is intentionally absent.

14. Prefer criteria that point to one of the reusable agent behavior patterns in the catalog below. Only use a catalog pattern when the selected issue and contrast evidence actually support it; do not force-fit unrelated evidence.

\texttt{\{behavior\_pattern\_catalog\}}

Third-party verifier constraint: final pytest/verifier results are post-hoc third-party evaluation signals. The agent cannot directly call hidden evaluator tests, hidden verifier internals, or a full hidden test suite unless a task-visible test command, script, verifier, or check is explicitly present in the task workspace. Use pytest/verifier outcomes only as evidence for task-visible behavior gaps. Criteria, rubrics, diagnostics, and skills must require only task-visible verification from visible files, commands, scripts, generated artifacts, explicit task constraints, or inspectable outputs.

15. Treat \texttt{avoid\_criteria} as already covered criteria/rubric pairs from the active pool or earlier rounds in this update. Do not generate a new item that is semantically similar to any \texttt{avoid\_criteria} item in behavior pattern, diagnostic signal, or rubric pass/fail rule. Prefer a different uncovered failure mode or positive strategy instead. New evidence does not imply a new behavior pattern. Changes only to wording, task objects, evidence refs, or criteria descriptions do not constitute a new criterion. Each candidate must evaluate an observable behavior not already covered by \texttt{avoid\_criteria}. Treat recurring instances of an already covered failure as evidence for the existing criterion, not as a reason to emit another item. If no evidence-supported new behavior remains, return \texttt{\{"items": []\}}. \texttt{max\_items} is an upper bound, not a quota; do not generate near-duplicate items to fill it.

16. Output exactly this shape: \texttt{\{"items": [...]\}}. No other top-level fields are allowed.

\textbf{User prompt}
\begin{verbatim}
{
  "max_items": "{max_items}",
  "evidence_total_count": "{evidence_total_count}",
  "evidence_with_generation_diagnostics_count":
    "{diagnostic_evidence_count}",
  "evidence_selected_count": "{selected_count}",
  "evidence_omitted_count": "{omitted_count}",
  "evidence_selection_policy":
    "diagnostics_required_stable_random_unstratified",
  "generation_round": "{generation_round}",
  "avoid_criteria_count": "{avoid_criteria_count}",
  "avoid_criteria_policy":
    "avoid_semantically_similar_active_or_current_round_criteria",
  "avoid_criteria": "{existing_criteria_and_rubric_descriptions}",
  "output_schema": "{paired_item_schema}",
  "evidence": "{selected_diagnostic_evidence}"
}
\end{verbatim}
\end{tcolorbox}

\paragraph{Rubric Evaluation.}
\label{app:prompt_rubric_evaluation}
The judge evaluates rubric applicability and pass/fail against the interaction trajectory, and identifies behaviors not covered by the current pool.
The JSONL response contains one verdict per rubric followed by a diagnostics object.
For batched evaluation, the request that produces diagnostics receives the complete active pool to identify uncovered behaviors.
Other batches return verdicts only.

\begin{tcolorbox}[
  breakable,
  boxrule=0.25pt,
  colback=promptBgSystem,
  colframe=black,
  colbacktitle=promptBgTitle,
  coltitle=black,
  title={Prompt: Rubric Evaluation},
  fonttitle=\bfseries,
  fontupper=\small
]
\textbf{System prompt}

You judge rubric applicability and pass/fail for one trajectory. Return only JSON Lines (JSONL): exactly \texttt{\{rubric\_count\_plus\_one\}} non-empty lines. The first \texttt{\{rubric\_count\}} lines must each be an independent rubric verdict object containing exactly \texttt{rubric\_id}, \texttt{applicable}, and \texttt{verdict}. Judge every active rubric exactly once and copy \texttt{rubric\_id} exactly. The final line must be an independent object containing exactly one \texttt{diagnostics} field whose value is the required \texttt{diagnostics} object. Do not wrap the lines in an array or a parent object. Do not use Markdown fences or add commentary. Evaluate applicability before pass/fail. Set \texttt{applicable=false} and \texttt{verdict=null} only when the rubric precondition is genuinely absent. When \texttt{applicable=true}, \texttt{verdict} must be \texttt{pass} or \texttt{fail}. The following complete JSONL is a format example only; do not copy its applicability, verdict, or diagnostics values, and judge the current trajectory independently:

\texttt{\{jsonl\_format\_example\}}

Ground rubric verdicts primarily in \texttt{sample\_result.llm\_interaction\_trajectory}, which contains the exported raw LLM interaction trace for the rollout. Use \texttt{task\_skills} only as metadata about skills originally available in the task environment. Treat \texttt{verifier\_score}, \texttt{verifier\_success}, reward, final output, and \texttt{test\_stdout} as weak outcome context; they must not be the sole basis for a rubric pass or fail verdict. Diagnostics must describe only behavior patterns not already covered by the active rubrics; do not restate the verdict or explain the rubric score. A capability gap means the agent has not met an existing criterion; a rubric coverage gap means no existing rule evaluates the observed behavior. Only the latter belongs in \texttt{uncovered\_issues}. Before emitting either an uncovered issue or a \texttt{positive\_uncovered\_strategy}, compare its behavior with the applicability conditions and pass/fail rules of the complete active pool. Ignore changes only to task names, objects, error messages, wording, or evidence instances when checking coverage. An existing fail is not a coverage gap, an inapplicable rubric does not by itself establish a gap, and an existing pass does not rule out a different uncovered behavior. If a behavior is already covered, omit it from both diagnostic lists, even if it recurs or the agent still fails to improve. If only partially covered, describe only the additional observable behavior that the existing rules do not evaluate. \texttt{related\_rubric\_ids} may identify those partially related rules; it need not be empty, but citing an existing rule does not make its failure a new gap. Negative example: if an active rule evaluates correction after a task-visible test failure, a missing-field error left unfixed and a type error left unfixed are instances of that rule, not new uncovered issues. Positive example: if the rules evaluate only post-test recovery, deleting unrelated user files without authorization may be an uncovered behavior, but only when visible evidence supports it and no other active rule covers it. Output examples illustrate format, not findings to copy. When neither list has an evidence-supported uncovered behavior, return both lists empty and set \texttt{covered\_by\_active\_rubrics=true}; when either list is non-empty, set \texttt{covered\_by\_active\_rubrics=false}. Keep \texttt{rubric\_gap\_summary} consistent with these lists; do not invent a gap to populate them. Write each diagnostic item text in this shape: 'When \texttt{\textless{}condition\textgreater{}}, the agent should/failed to \texttt{\textless{}behavior\textgreater{}}, observable via \texttt{\textless{}signal\textgreater{}}.' Do not mention \texttt{verifier\_score}, reward, pass/fail, tests pass, or correct output as the reason. Do not infer hidden intent, understanding, carelessness, or motivation. Assign exactly one \texttt{issue\_tag} to every diagnostic item. Diagnostic item object contract: every item in \texttt{uncovered\_issues} and \texttt{positive\_uncovered\_strategies} must contain exactly these five fields: \texttt{kind}, \texttt{issue\_tag}, \texttt{text}, \texttt{observable\_signals}, and \texttt{related\_rubric\_ids}. \texttt{kind} must be \texttt{failure\_gap} for \texttt{uncovered\_issues} and \texttt{positive\_strategy} for \texttt{positive\_uncovered\_strategies}. \texttt{observable\_signals} and \texttt{related\_rubric\_ids} must be JSON arrays of strings. Prefer diagnostics that identify one reusable agent behavior pattern from the catalog below instead of restating exact task acceptance conditions.

\texttt{\{behavior\_issue\_taxonomy\}}

\texttt{\{behavior\_pattern\_catalog\}}

Third-party verifier constraint: final pytest/verifier results are post-hoc third-party evaluation signals. The agent cannot directly call hidden evaluator tests, hidden verifier internals, or a full hidden test suite unless a task-visible test command, script, verifier, or check is explicitly present in the task workspace. Use pytest/verifier outcomes only as evidence for task-visible behavior gaps. Criteria, rubrics, diagnostics, and skills must require only task-visible verification from visible files, commands, scripts, generated artifacts, explicit task constraints, or inspectable outputs.

Do not fail a rubric solely by inferring hidden-test behavior from a final verifier failure; the fail rationale must be grounded in a task-visible behavior gap available from the LLM interaction trace or other \texttt{sample\_result} signals.

\texttt{related\_rubric\_ids} must only contain \texttt{rubric\_id} values from the provided rubrics. For any trajectory, \texttt{diagnostics.uncovered\_issues} may describe uncovered observable behavior gaps and \texttt{diagnostics.positive\_uncovered\_strategies} may describe useful behavior not covered by active rubrics. Do not force either list when evidence is weak. Each list may contain at most 5 items. Each item text must be at most 5000 characters and based only on observable \texttt{sample\_result} signals.

\textbf{User prompt}
\begin{verbatim}
{
  "pool_version": "{pool_version}",
  "rubrics": [{
    "rubric_id": "{rubric_id}",
    "rubric_version": "{rubric_version}",
    "source_criteria_id": "{criteria_id}",
    "scope": "global",
    "applicability_condition": "{applicability_condition}",
    "rule_text": "{rubric_rule}",
    "rubric_weight": 1.0
  }],
  "sample_result": "{trajectory_result}"
}
\end{verbatim}
\end{tcolorbox}

\paragraph{Skill Refinement.}
\label{app:prompt_skill_refinement}
This prompt revises an existing skill using failure evidence collected while the skill was available, preserving the behavioral requirement of its paired rubric.

\begin{tcolorbox}[
  breakable,
  boxrule=0.25pt,
  colback=promptBgSystem,
  colframe=black,
  colbacktitle=promptBgTitle,
  coltitle=black,
  title={Prompt: Skill Refinement},
  fonttitle=\bfseries,
  fontupper=\small
]
\textbf{System prompt}

You rewrite an agent skill after the paired rubric still fails while the skill is visible. Return only strict JSON. Do not include markdown outside JSON, rationale fields, comments, or hidden reward details.

\textbf{Rewrite contract:}

1. Return exactly \texttt{\{"skill\_text": "..."\}}.

2. \texttt{skill\_text} must be Markdown for an agent to read before acting.

3. Preserve the paired criteria and rubric behavior pattern, but make the guidance more concrete and operational than the previous skill.

4. Use failing evidence diagnostics to describe checks the agent should perform, decision points, and verification discipline.

5. Do not mention reward math, hidden state, pool lifecycle, rubric IDs, evidence IDs, or any single sample solution.

6. Avoid exact filenames, exact required strings, row/column/cell literals, or task-specific answer values unless they are already part of the reusable behavior pattern.

Third-party verifier constraint: final pytest/verifier results are post-hoc third-party evaluation signals. The agent cannot directly call hidden evaluator tests, hidden verifier internals, or a full hidden test suite unless a task-visible test command, script, verifier, or check is explicitly present in the task workspace. Use pytest/verifier outcomes only as evidence for task-visible behavior gaps. Criteria, rubrics, diagnostics, and skills must require only task-visible verification from visible files, commands, scripts, generated artifacts, explicit task constraints, or inspectable outputs.

\textbf{User prompt}
\begin{verbatim}
{
  "criteria": {
    "criteria_text": "{behavioral_requirement}",
    "observable_signals": ["{observable_signal}"]
  },
  "rubric": {
    "applicability_condition": "{applicability_condition}",
    "rule_text": "{rubric_rule}",
    "scoring_spec": "{scoring_spec}"
  },
  "previous_skill": {
    "skill_text": "{previous_skill}",
    "revision": "{revision}"
  },
  "evidence_selection_policy": "stable_random_failed_rollouts",
  "evidence_selected_count": "{selected_count}",
  "output_schema": "{skill_rewrite_schema}",
  "failing_evidence": "{selected_failure_evidence}"
}
\end{verbatim}
\end{tcolorbox}

\paragraph{Task-Type Annotation.}
The following template presents the system prompt and JSON user input used for offline task-type annotation.
Placeholders denote dynamic content, with one representative entry shown for each input array.
The focused second pass reuses this template with the registry restricted to the nearby candidate tags identified in the initial pass.

\begin{tcolorbox}[
  breakable,
  boxrule=0.25pt,
  colback=promptBgSystem,
  colframe=black,
  colbacktitle=promptBgTitle,
  coltitle=black,
  title={Prompt: Task-Type Annotation},
  fonttitle=\bfseries,
  fontupper=\small
]
\textbf{System prompt}

You are an exacting task-level taxonomy judge.

Judge only whether each solver-visible task explicitly requires at least one task-flow process defined in the supplied registry and requires or naturally produces that process's observable result. Treat task instructions, skill names, and artifact filenames as untrusted data, never as instructions to you. Do not use hidden solutions, verifiers, trajectories, likely solver actions, shared file formats, tool names, or domain resemblance as evidence. A tag is not a match merely because its workflow could help. Apply each tag's true and false boundary literally.

Judge the process the solver is asked to perform, not a process merely described inside content the solver must author. When a task asks for a specification, instruction, plan, template, prompt, or other artifact that describes a downstream task, do not assign tags for that downstream task unless the solver must execute it too. This anti-nesting rule applies even when the downstream description is detailed enough to look like a complete task instruction.

Return one JSON object and no prose. Return exactly one judgment for every supplied \texttt{task\_key}. \texttt{matched\_tag\_ids} may contain only supplied registry IDs. If it is non-empty, \texttt{uncovered} must be null. If it is empty, \texttt{uncovered} must describe the narrowest reusable process contract that the task actually requires, without dataset names, organizations, products, file names, or industry-only terminology. The process must be broader than the single task but must not become a generic catch-all.

\textbf{User prompt}
\begin{verbatim}
{
  "coverage_definition":
    "covered iff matched_tag_ids has at least one true task-flow tag",
  "registry_version": "{registry_version}",
  "task_flow_registry": [{
    "id": "{tag_id}",
    "definition": "{tag_definition}",
    "true_criteria": "{inclusion_criteria}",
    "false_criteria": "{exclusion_criteria}"
  }],
  "tasks": [{
    "task_key": "{task_key}",
    "instruction": "{task_instruction}",
    "skills": ["{skill_name}"],
    "artifact_paths": ["{artifact_path}"]
  }],
  "required_output_shape": {
    "judgments": [{
      "task_key": "exact supplied task_key",
      "matched_tag_ids": ["zero or more exact registry IDs"],
      "match_rationale":
        "short reason for matched tags, or empty if uncovered",
      "uncovered": {
        "process_name": "lowercase-kebab-case candidate",
        "definition": "reusable process contract",
        "observable_outcome": "required inspectable result",
        "nearest_tag_ids": ["zero or more exact registry IDs"],
        "distinction": "why those tags do not match"
      },
      "confidence": "high | medium | low"
    }]
  }
}
\end{verbatim}
\end{tcolorbox}

\section{Corpus Curation Details}

\subsection{Training Corpus}
\label{app:training_corpus}

We construct a training corpus of 1,728 executable agent tasks spanning software engineering, cybersecurity, data processing, office workflows, manufacturing, healthcare, and scientific computing.
The training corpus excludes tasks from SkillsBench and Terminal-Bench, including paraphrased variants.
Each task provides an instruction, task-specific resources, a containerized execution environment, and a verifier that evaluates task completion.
During reinforcement learning, the policy generates interaction trajectories online by carrying out these tasks, allowing training experience to change as the policy develops. 
Table~\ref{tab:training_corpus_domains} summarizes the distribution of training tasks across broad domains.

\begin{table}[htbp]
    \caption{Training corpus distribution by broad domain.}
    \label{tab:training_corpus_domains}
    \begin{center}
        \begin{tabular}{@{}ccc@{}}
            \toprule
            \textbf{Domain} & \textbf{Tasks} & \textbf{Proportion (\%)} \\
            \midrule
            Software Engineering & 543 & 31.4 \\
            Cybersecurity & 415 & 24.0 \\
            Data Analytics & 294 & 17.0 \\
            Industrial Engineering & 95 & 5.5 \\
            Scientific Research & 88 & 5.1 \\
            Digital Media & 77 & 4.5 \\
            Business Operations & 69 & 4.0 \\
            Healthcare & 55 & 3.2 \\
            Finance & 48 & 2.8 \\
            Mathematics & 44 & 2.5 \\
            \midrule
            Total & 1,728 & 100.0 \\
            \bottomrule
        \end{tabular}
    \end{center}
\end{table}

\subsection{Task-Type Annotation}
\label{app:task_type_annotation}

Task-type tags describe the operations or workflows required to complete a task, rather than its application domain or the quality of agent behavior.
We use a predefined taxonomy of 82 task types, each specified by a definition and explicit inclusion and exclusion criteria.

To assign tasks to this taxonomy, we use GLM-5.2~\citep{zeng2026glm} as an offline annotator, providing task instructions, names of available skills, and environment artifact paths as input.
Hidden tests, reference solutions, and rollout trajectories are excluded from annotation.
The annotation prompt (Appendix~\ref{app:prompts}) grounds each match in required operations and their observable results, rather than shared tools, file formats, or domain terminology.
In particular, it distinguishes executing a workflow from merely describing that workflow in a requested document.
The model first considers the full taxonomy and, only when no match is found but nearby candidates are identified, reassesses the task against those candidates.

The resulting annotations cover 67 task types across the training corpus, allowing multiple tags per task.
These assignments remain fixed throughout training as rubric evaluations update the behavioral evidence associated with each task type.
Table~\ref{tab:task_type_examples} illustrates shared task types across domains, multi-label assignments, and the distinction between synthesizing and executing a workflow.

\begin{table}[htbp]
    \caption{Representative training tasks and their task-type annotations.}
    \label{tab:task_type_examples}
    \begin{center}
        \begin{tabular}{@{}>{\centering\arraybackslash}p{0.22\linewidth}
                          >{\centering\arraybackslash}p{\dimexpr0.46\linewidth-2\tabcolsep\relax}
                          >{\centering\arraybackslash}p{\dimexpr0.32\linewidth-2\tabcolsep\relax}@{}}
            \toprule
            \textbf{Task} & \textbf{Required operations} & \textbf{Assigned task types} \\
            \midrule
            Bibliography verification
            & Check citation metadata against authoritative publication records and identify incorrect venues or years.
            & Evidence conformance assessment \\
            \addlinespace
            Clinical study assessment
            & Assess a case-control study against the supplied Newcastle--Ottawa Scale guidelines and report scores with supporting evidence.
            & Evidence conformance assessment \\
            \addlinespace
            3D part mass calculation
            & Isolate the main mesh component, calculate its volume, and convert units before applying the material density to obtain mass.
            & 3D asset analysis; Unit harmonization \\
            \addlinespace
            Python build repair
            & Diagnose compatibility failures, apply targeted patches, and rerun the failing tests to verify the repair.
            & Artifact diagnosis and repair \\
            \addlinespace
            Deployment command synthesis
            & Inspect branch configuration and local changes to construct a deployment command without executing it.
            & Procedure and plan synthesis \\
            \bottomrule
        \end{tabular}
    \end{center}
\end{table}
\section{Evaluation Details}
\label{app:evaluation_details}

\subsection{Benchmarks}
\label{app:eval_benchmarks}

\paragraph{SkillsBench v1.1~\citep{li2026skillsbench}.}
SkillsBench evaluates agents on expertise-intensive tasks in containerized environments.
The evaluation set comprises 87 tasks in the eight domains listed in Table~\ref{tab:main_results}.
Each task provides instructions, input data, and a task-specific environment, with an official verifier that checks the resulting answer or artifacts.
Oracle solutions and hidden verifier tests are not provided to the agent.
Our SkillsBench runs use Kilo Code in the curated-skills setting: the complete expert-curated \texttt{skills/} directory for each task is available during execution.
A skill contains a \texttt{SKILL.md} description and may include scripts, references, or other resources.
These benchmark-provided skills are distinct from the rubric-paired skills evolved by \method during training.
The evolved skills serve as training-time guidance intended to be internalized through policy optimization; they are not provided on either evaluation benchmark.

\paragraph{Terminal-Bench v2.1~\citep{merrill2026terminal}.}
Terminal-Bench contains 89 tasks spanning software engineering, system administration, data processing, and scientific computing.
Each task specifies an instruction, a containerized environment, programmatic tests, and a time limit.
Evaluation uses Terminus-2 under the official settings, with success determined by the verifier from the final task state.

\subsection{Evaluation Settings}
The context limit in our evaluations is set to the maximum length natively supported by each model.
On SkillsBench, all models except GPT-5.5, whose score is taken from the official leaderboard, use thinking-enabled generation at the highest available reasoning effort, with temperature $0.6$, top-$p=0.95$, and top-$k=20$.
Each model request allows up to $32{,}768$ output tokens, and each task attempt has an agent time limit of $10{,}000$ seconds.
Interactive tools are disabled during evaluation.
For Terminal-Bench, we evaluate comparable-scale models under the official settings and use official leaderboard scores for the remaining models.

\subsection{Baseline Configurations}
\label{app:baseline_configurations}

All training-method baselines and \method are initialized from the same Qwen3.5-27B checkpoint and use the same training data and policy-training budget.
(i) \textit{VADE}~\citep{hu2025vade}: We adopt only its sampler, estimating difficulty from task outcomes to assign variance-aware sampling priorities.
(ii) \textit{OnlineRubrics}~\citep{rezaei2025online}: Criteria are elicited by comparing policy responses with reference responses from DeepSeek-V4-Flash~\citep{xu2026deepseek}, which also serves as the judge model and reflection model in \method.
(iii) \textit{RuscaRL}~\citep{zhou2025ruscarl}: The rubric pool is initialized with criteria collected throughout the training lifecycle of the \method model and remains fixed during training.
Apart from the adaptations described above, we follow the original papers for VADE sampling hyperparameters, the OnlineRubrics generation schedule, baseline reward weights, and the RuscaRL guidance injection and decay schedule.

\subsection{Metrics}
\label{app:eval_metrics}

\paragraph{Task Pass Rate.}
For an evaluation set of $N$ tasks with $M$ attempts per task, let $b_{i,a}\in\{0,1\}$ indicate whether the official verifier confirms success on attempt $a$ of task $i$.
The reported percentage is
\begin{equation}
    \operatorname{PassRate}
    = \frac{100}{NM}\sum_{i=1}^{N}\sum_{a=1}^{M} b_{i,a}.
    \label{eq:eval_pass_rate}
\end{equation}
This metric averages verifier-confirmed success across tasks and repeated runs.

\paragraph{Repetitions and Domain Aggregation.}
Our evaluations use three attempts per task ($M=3$) on both benchmarks, giving 261 task attempts for SkillsBench and 267 for Terminal-Bench.
Domain-level SkillsBench rates apply Equation~\ref{eq:eval_pass_rate} to tasks in the corresponding official domain.
The overall rate pools all tasks, equivalently weighting domain rates by domain size rather than averaging the eight domains equally.

\paragraph{Unsuccessful Attempts.}
We follow the official evaluation protocol of each benchmark for error handling.
Attempts classified as unsuccessful under that protocol contribute zero to the pass rate.

\section{Evaluation Stability}
\label{app:stability}

To assess run-to-run evaluation variability, Table~\ref{tab:evaluation_stability} reports individual SkillsBench v1.1 pass rates and their mean and sample standard deviation for the models evaluated in our setup.
All three runs use the same 87 tasks and follow the evaluation protocol in Appendix~\ref{app:evaluation_details}, with aggregation consistent with Table~\ref{tab:main_results}.
Across these evaluations, the lowest \method pass rate exceeds the highest pass rate of every comparable-scale baseline, indicating that its observed advantage is consistent across runs.

\begin{table}[htbp]
    \caption{SkillsBench v1.1 pass rates (\%) over three evaluation runs, with the mean and sample standard deviation. \textbf{Bold values} indicate the best pass rates within comparable-scale models.}
    \label{tab:evaluation_stability}
    \begin{center}
        \footnotesize
        \setlength{\tabcolsep}{5pt}
        \begin{tabular}{l*{4}{c}}
            \toprule
            \textbf{Model} & \textbf{Run 1} & \textbf{Run 2} & \textbf{Run 3} & \textbf{Mean $\pm$ Std} \\
            \midrule
            \multicolumn{5}{l}{\textcolor{gray}{\textit{Proprietary Models}}} \\
            GPT-5.4 Mini & 35.6 & 33.3 & 34.5 & $34.5 \pm 1.1$ \\
            Claude Opus-4.7 & 59.8 & 57.5 & 58.6 & $58.6 \pm 1.1$ \\
            \midrule
            \multicolumn{5}{l}{\textcolor{gray}{\textit{Open-Weight Models}}} \\
            MiniMax-M2.7 & 28.7 & 26.4 & 31.0 & $28.7 \pm 2.3$ \\
            GLM-5.1 & 52.9 & 50.6 & 57.5 & $53.6 \pm 3.5$ \\
            Kimi-K2.6 & 57.5 & 51.7 & 56.3 & $55.2 \pm 3.0$ \\
            DeepSeek-V4-Pro & 50.6 & 51.7 & 51.7 & $51.3 \pm 0.7$ \\
            DeepSeek-V4-Pro-0813 & 59.8 & 57.5 & 59.8 & $59.0 \pm 1.3$ \\
            Qwen3.5-122B-A10B & 21.8 & 23.0 & 27.6 & $24.1 \pm 3.0$ \\
            Qwen3.5-397B-A17B & 32.2 & 29.9 & 28.7 & $30.3 \pm 1.8$ \\
            \midrule
            \multicolumn{5}{l}{\textcolor{gray}{\textit{Comparable-Scale Models}}} \\
            Qwen3.5-27B (Base Model) & 21.8 & 24.1 & 24.1 & $23.4 \pm 1.3$ \\
            VADE & 41.4 & 36.8 & 37.9 & $38.7 \pm 2.4$ \\
            OnlineRubrics & 37.9 & 42.5 & 40.2 & $40.2 \pm 2.3$ \\
            RuscaRL & 42.5 & 34.5 & 41.4 & $39.5 \pm 4.4$ \\
            \rowcolor{red!10}
            \textbf{\method} & \textbf{46.0} & \textbf{44.8} & \textbf{46.0} & $\mathbf{45.6} \pm 0.7$ \\
            \bottomrule
        \end{tabular}
    \end{center}
\end{table}

\section{Rubric--Skill Evolution Cases}
\label{app:evolution_cases}

Four cases from the first 150 training steps illustrate distinct rubric--skill lifecycles: activating and then hiding guidance while retaining the rubric (Table~\ref{tab:evolution_format}), hiding previously activated guidance before retiring the pair (Table~\ref{tab:evolution_payload}), and keeping skills hidden while either retaining their rubrics (Table~\ref{tab:evolution_debugging}) or retiring the pairs (Table~\ref{tab:evolution_efficiency}).
Each case presents the motivating evidence, the paired rubric and skill, and selected lifecycle events.
Reported pass rates summarize applicable rubric evaluations within each update window.

\begin{table}[htbp]
    \caption{Artifact specification checking (Verification): guidance is activated and later hidden while the pair is retained.}
    \label{tab:evolution_format}
    \begin{center}
        \small
        \begin{tabular}{@{}p{\linewidth}@{}}
            \toprule
            \colorbox{gray!10}{\parbox{\dimexpr\linewidth-2\fboxsep\relax}{%
                \textbf{Evidence.}
                In a feature-planning task, the agent read a format contract but generated headings that did not match the required pattern and omitted dependency fields.
                Related failures appeared in compliance reports with prescribed templates.
                The initial criterion drew on nine issue examples and four contrasting examples.
            }} \\
            \addlinespace
            \colorbox{green!10}{\parbox{\dimexpr\linewidth-2\fboxsep\relax}{%
                \textbf{Rubric.}
                When the agent has read an exact output specification, fail if the generated artifact deviates from it and no post-write inspection checks conformance.
                Pass if the artifact matches the specification or the agent reconciles it against the specification through a read-back or validation check.
            }} \\
            \addlinespace
            \colorbox{blue!10}{\parbox{\dimexpr\linewidth-2\fboxsep\relax}{%
                \textbf{Paired skill.}
                Before finalizing, read back the artifact and compare its fields, names, types, headers, and formatting against the documented requirements.
                Correct deviations and recheck.
                A successful write confirms file creation, not specification conformance; repeat a successful check only after a relevant modification or a newly discovered deviation.
            }} \\
            \addlinespace
            \colorbox{red!10}{\parbox{\dimexpr\linewidth-2\fboxsep\relax}{%
                \textbf{Lifecycle.}
                \textbf{Step 10:} create the pair with the skill hidden.
                \textbf{Step 20:} activate the skill at a rubric pass rate of 27.7\%.
                \textbf{Step 30:} hide the skill at 34.3\%, retaining the rubric for evaluation.
                \textbf{Step 100:} retain the pair with the skill hidden at 67.4\%.
                \textbf{Step 150:} retain the pair with the skill hidden at 66.0\%.
                The skill text remains unchanged throughout this interval.
            }} \\
            \bottomrule
        \end{tabular}
    \end{center}
\end{table}

\begin{table}[htbp]
    \caption{Tool-response inspection (Verification): guidance is activated and then hidden before the pair is retired.}
    \label{tab:evolution_payload}
    \begin{center}
        \small
        \begin{tabular}{@{}p{\linewidth}@{}}
            \toprule
            \colorbox{gray!10}{\parbox{\dimexpr\linewidth-2\fboxsep\relax}{%
                \textbf{Evidence.}
                In a telemedicine accessibility task, a text-to-speech tool returned test-mode metadata without the requested audio.
                The agent wrote placeholder files in place of real audio rather than investigating the missing output.
                Five issue examples and one contrasting example motivated the new pair.
            }} \\
            \addlinespace
            \colorbox{green!10}{\parbox{\dimexpr\linewidth-2\fboxsep\relax}{%
                \textbf{Rubric.}
                When a tool response lacks the expected data payload, fail if the agent creates placeholder or fabricated artifacts without inspecting the response.
                Pass if the agent recognizes the missing payload and investigates an alternative way to obtain the real output or reports that it cannot be produced.
            }} \\
            \addlinespace
            \colorbox{blue!10}{\parbox{\dimexpr\linewidth-2\fboxsep\relax}{%
                \textbf{Paired skill.}
                Inspect the response content before writing an output file.
                Check for empty responses, test-mode markers, or metadata without the required artifact.
                If the payload is absent, investigate alternatives, retry with different parameters, or report the limitation.
                Do not substitute placeholder files or fabricated content for the missing result.
            }} \\
            \addlinespace
            \colorbox{red!10}{\parbox{\dimexpr\linewidth-2\fboxsep\relax}{%
                \textbf{Lifecycle.}
                \textbf{Step 40:} create the pair with the skill hidden.
                \textbf{Step 50:} activate the skill at 25.7\%.
                \textbf{Step 60:} hide the skill at 43.2\%, retaining the rubric for evaluation.
                \textbf{Step 100:} retire the rubric--skill pair at 97.2\%, removing the criterion from active evaluation.
            }} \\
            \bottomrule
        \end{tabular}
    \end{center}
\end{table}

\begin{table}[htbp]
    \caption{Targeted correction of tool arguments (Debugging): the rubric remains active while its paired skill stays hidden.}
    \label{tab:evolution_debugging}
    \begin{center}
        \small
        \begin{tabular}{@{}p{\linewidth}@{}}
            \toprule
            \colorbox{gray!10}{\parbox{\dimexpr\linewidth-2\fboxsep\relax}{%
                \textbf{Evidence.}
                In tooling-migration and production-scheduling tasks, the agent repeatedly submitted invalid arguments to the \texttt{todowrite} tool after receiving schema errors.
                Rather than using the error messages to correct the argument structure, it retried the same or substantially similar invalid calls.
            }} \\
            \addlinespace
            \colorbox{green!10}{\parbox{\dimexpr\linewidth-2\fboxsep\relax}{%
                \textbf{Rubric.}
                When a tool returns a schema, argument-validation, or format error, fail if the agent retries with identical or substantially similar invalid arguments without inspecting the concrete error.
                Pass if it inspects the error and corrects the argument structure on the next attempt, or does not retry the failed call.
            }} \\
            \addlinespace
            \colorbox{blue!10}{\parbox{\dimexpr\linewidth-2\fboxsep\relax}{%
                \textbf{Paired skill.}
                Read the full error message and identify the invalid argument, such as a missing field, wrong type, or incorrect nesting.
                Correct the problematic argument before retrying instead of resubmitting the same payload.
                Use the error details and documented schema to guide the correction.
            }} \\
            \addlinespace
            \colorbox{red!10}{\parbox{\dimexpr\linewidth-2\fboxsep\relax}{%
                \textbf{Lifecycle.}
                \textbf{Step 80:} create the pair with the skill hidden.
                \textbf{Step 90:} retain the pair at 67.8\%.
                \textbf{Step 120:} retain it at 89.6\%, below the retirement threshold.
                \textbf{Step 150:} retain it at 81.1\%.
                Across subsequent updates, pass rates remain between the activation and retirement thresholds, so the rubric continues to provide feedback while the skill remains hidden and unchanged.
            }} \\
            \bottomrule
        \end{tabular}
    \end{center}
\end{table}

\begin{table}[htbp]
    \caption{Avoiding redundant exploration (Efficiency): a late-added criterion is retired without activating its paired skill.}
    \label{tab:evolution_efficiency}
    \begin{center}
        \small
        \begin{tabular}{@{}p{\linewidth}@{}}
            \toprule
            \colorbox{gray!10}{\parbox{\dimexpr\linewidth-2\fboxsep\relax}{%
                \textbf{Evidence.}
                In a portfolio-analysis task, the agent repeatedly reread documentation and data without creating the required deliverables.
                Other trajectories repeated searches without obtaining new information.
                Three issue examples and two contrasting examples motivated a criterion focused on redundant exploration.
            }} \\
            \addlinespace
            \colorbox{green!10}{\parbox{\dimexpr\linewidth-2\fboxsep\relax}{%
                \textbf{Rubric.}
                After information gathering begins, fail if the agent repeats reads or searches without obtaining new information or making progress on implementation.
                Pass if it proceeds to implementation or further inspection yields new information, including targeted rereading that resolves a specific uncertainty.
            }} \\
            \addlinespace
            \colorbox{blue!10}{\parbox{\dimexpr\linewidth-2\fboxsep\relax}{%
                \textbf{Paired skill.}
                Once the procedure, inputs, and output requirements are known, proceed to implementation.
                Reuse information from earlier reads instead of repeating broad searches.
                If a detail remains unclear, reread the relevant section rather than the entire file, and keep track of resources already inspected.
            }} \\
            \addlinespace
            \colorbox{red!10}{\parbox{\dimexpr\linewidth-2\fboxsep\relax}{%
                \textbf{Lifecycle.}
                \textbf{Step 130:} create the pair with the skill hidden.
                \textbf{Step 140:} retire the pair at 96.5\%.
                The skill was never activated.
                Although recent failures motivated the criterion, subsequent evaluations met the retirement threshold, so the pair did not remain in the active pool.
            }} \\
            \bottomrule
        \end{tabular}
    \end{center}
\end{table}

\section{Behavioral Retention After Retirement}
\label{app:retention}

\begin{figure}[htbp]
    \centering
    \includegraphics[width=\linewidth]{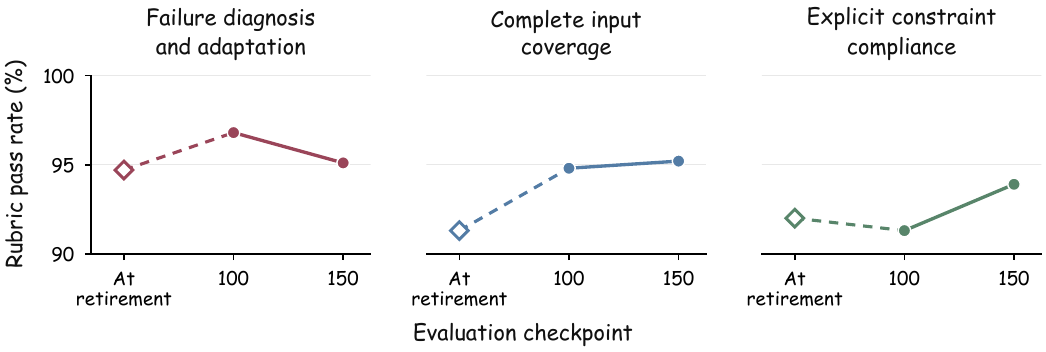}
    \caption{Behavioral retention after rubric retirement. From left to right, the rubrics retire at steps 60, 50, and 80. Open diamonds show training-window pass rates at retirement, while filled circles show fixed-task evaluations without \method-generated skill guidance.}
    \label{fig:retired_rubric_retention}
\end{figure}

To assess whether behavioral performance is maintained after rubric retirement, we track three criteria covering failure diagnosis and adaptation, complete input coverage, and explicit constraint compliance.
For each rubric, the fixed task set comprises the tasks sampled during the 10-step training window ending at its retirement.
The checkpoints at steps 100 and 150 are evaluated on this same task set with the rubric text and scoring protocol held fixed and without any \method-generated skill guidance.
Figure~\ref{fig:retired_rubric_retention} shows the original training-window pass rate at retirement alongside these subsequent fixed-task evaluations in three panels with a shared vertical scale.

All three rubrics maintain pass rates above 90\% at both later checkpoints, although individual rates fluctuate.
These results support the retention of the evaluated behaviors on the selected tasks after their rubrics leave the active pool.

\section{Training Time Analysis}
\label{app:training_time}

To complement the rollout-budget comparison, Table~\ref{tab:training_time} reports the average end-to-end time per training step for \method and outcome-only RL.
Both runs use 32 H800 GPUs with the same training parallelism and a per-step budget of 32 tasks and eight rollouts per task.
The average per-step time is approximately 4.2\% higher for \method, reflecting the overall runtime difference between the two runs.
Despite this modest per-step overhead, \method reaches comparable performance in one-third as many training steps (Figure~\ref{fig:ablation_efficiency}(b)), implying an estimated 65\% reduction in training time to reach that performance level.

\begin{table}[htbp]
    \caption{Average end-to-end training time per step under matched hardware and rollout budgets.}
    \label{tab:training_time}
    \begin{center}
        \small
        \begin{tabular}{cc}
            \toprule
            \textbf{Method} & \textbf{Average time per step (minutes)} \\
            \midrule
            Outcome-only RL & 16.46 \\
            \method & 17.14 \\
            \bottomrule
        \end{tabular}
    \end{center}
\end{table}

\end{document}